\documentclass[11pt]{article}

\usepackage[preprint]{acl}

\usepackage{longtable}
\usepackage{mathtools} 
\usepackage{times}
\usepackage{latexsym}
\usepackage{booktabs}
\usepackage{tabularx}
\usepackage{multirow} 
\usepackage{graphicx} 
\usepackage{wrapfig}  
\usepackage{caption}
\usepackage{subcaption}
\usepackage{hyperref}
\usepackage{url}
\usepackage{ulem}
\usepackage{pifont}
\usepackage{xcolor}

\definecolor{darkgreen}{rgb}{0.0, 0.5, 0.0}
\definecolor{darkred}{rgb}{0.8, 0.0, 0.0}

\newcommand{\up}[1]{{\scriptsize\color{darkgreen}{$\uparrow$#1}}}
\newcommand{\down}[1]{{\scriptsize\color{darkred}{$\downarrow$#1}}}
\newcommand{\neut}[1]{{\scriptsize\color{gray}{#1}}}

\newcommand{\cmark}{\textcolor{green!70!black}{\ding{51}}} 
\newcommand{\xmark}{\textcolor{red!70!black}{\ding{55}}}    
\usepackage{paralist}
\usepackage{cuted}   
\usepackage{amsmath,amsfonts,bm}

\usepackage{amsmath,amsfonts,bm}

\def\eqref#1{equation~\ref{#1}}

\def\1{\bm{1}}

\DeclareMathAlphabet{\mathsfit}{\encodingdefault}{\sfdefault}{m}{sl}
\SetMathAlphabet{\mathsfit}{bold}{\encodingdefault}{\sfdefault}{bx}{n}

\newcommand{\method}[1]{\textsc{RLAR}}
\usepackage[table,xcdraw]{xcolor}
\usepackage[breakable,skins]{tcolorbox}
\usepackage{listings}
\lstdefinestyle{mypython}{
    language=Python,
    basicstyle=\ttfamily\footnotesize,
    keywordstyle=\color{blue},
    stringstyle=\color{red},
    commentstyle=\color{green!50!black},
    showstringspaces=false,
    breaklines=true,        
    breakatwhitespace=false, 
    columns=fullflexible,
    numbers=left,
    numberstyle=\tiny,
    frame=single,
    tabsize=4
}
\definecolor{lightgreen}{HTML}{D9EAD3}
\usepackage{tikz}
\usetikzlibrary{shapes.geometric, arrows, positioning, calc}
\tikzstyle{startstop} = [rectangle, rounded corners, minimum width=2.5cm, minimum height=1cm,text centered, draw=black, fill=red!20]
\tikzstyle{process} = [rectangle, minimum width=3cm, minimum height=1cm, text centered, draw=black, fill=blue!20]
\tikzstyle{decision} = [diamond, aspect=2, text centered, draw=black, fill=green!20]
\tikzstyle{data} = [parallelogram, minimum width=2.5cm, minimum height=1cm, text centered, draw=black, fill=orange!20]
\tikzstyle{arrow} = [thick,->,>=stealth]
\tikzstyle{db} = [cylinder, shape border rotate=90, aspect=0.5, draw, fill=yellow!20, minimum height=1cm,minimum width=1.5cm]

\tcbset{
    promptbox/.style={
        breakable,
        colframe=blue!50!black,
        colback=blue!5,
        fonttitle=\bfseries,
        left=2mm,
        right=2mm,
        top=2mm,
        bottom=2mm,
        before skip=10pt,
        after skip=10pt
    }
}

\usepackage[T1]{fontenc}

\usepackage[utf8]{inputenc}

\usepackage{microtype}

\usepackage{inconsolata}

\usepackage{graphicx}

\title{RLAR: An Agentic Reward System for Multi-task Reinforcement Learning on Large Language Models}

\author{
    Andrew Zhuoer Feng$^{\ast, \alpha}$, 
    Cunxiang Wang$^{\ast, \beta, \alpha}$,
    Bosi Wen$^{\alpha}$,
    Yidong Wang$^{\gamma}$, \\
    \bf Yu Luo$^{\alpha}$,
    Hongning Wang$^{\alpha}$,
    Minlie Huang$^{\alpha}$ \\
    $^\alpha$Tsinghua Univeristy\quad $^\beta$Z.ai\quad $^\gamma$Peking University\\
    \texttt{fze22, aihuang@tsinghua.edu.cn} \quad \texttt{wangcunxiang303@gmail.com}
}

\begin{document}
\maketitle

{
    \let\thefootnote\relax\footnotetext{
    \begin{tabular}{@{}l@{\hspace{0.5em}}l@{}}
        $^\ast$ & Contributed equally to this work. \\
        $^\dagger$ & Work done when A. Z. Feng interned at Z.ai.
    \end{tabular}
    }
}

\begin{abstract}
Large language model alignment via reinforcement learning depends critically on reward function quality. 
However, static, domain-specific reward models are often costly to train and exhibit poor generalization in out-of-distribution scenarios encountered during RL iterations.
We present \method{} (\underline{\textbf{R}}einforcement \underline{\textbf{L}}earning from \underline{\textbf{A}}gent \underline{\textbf{R}}ewards), an agent‑driven framework that dynamically assigns tailored reward functions to individual queries.
Specifically, \method{} transforms reward acquisition into a dynamic tool synthesis and invocation task.
It leverages LLM agents to autonomously retrieve optimal reward models from the Internet and synthesize programmatic verifiers through code generation.
This allows the reward system to self-evolve with the shifting data distributions during training.
Experimental results demonstrate that RLAR yields consistent performance gains ranging from \(10\%\) to \(60\%\) across mathematics, coding, translation, and dialogue tasks.
On \textsc{RewardBench-V2}, \method{} significantly outperforms static baselines and approaches the performance upper bound, demonstrating superior generalization through dynamic reward orchestration.\footnote{The data and code are available on \url{https://github.com/ZhuoerFeng/RLAR}.}
\end{abstract}

\section{Introduction}

Large language model (LLM) alignment via reinforcement learning (RL) has achieved substantial progress, where a policy model's parameters are iteratively updated to maximize rewards from an oracle~\citep{ppo, ouyangrlhf, grpo}. The effectiveness of this process hinges on the quality of the reward system, which is required to reliably score candidate responses in the context of queries and reference responses. This requirement often dictates the ceiling of model performance, thus becoming a core focus.

\begin{figure}
    \includegraphics[width=1.0\linewidth]{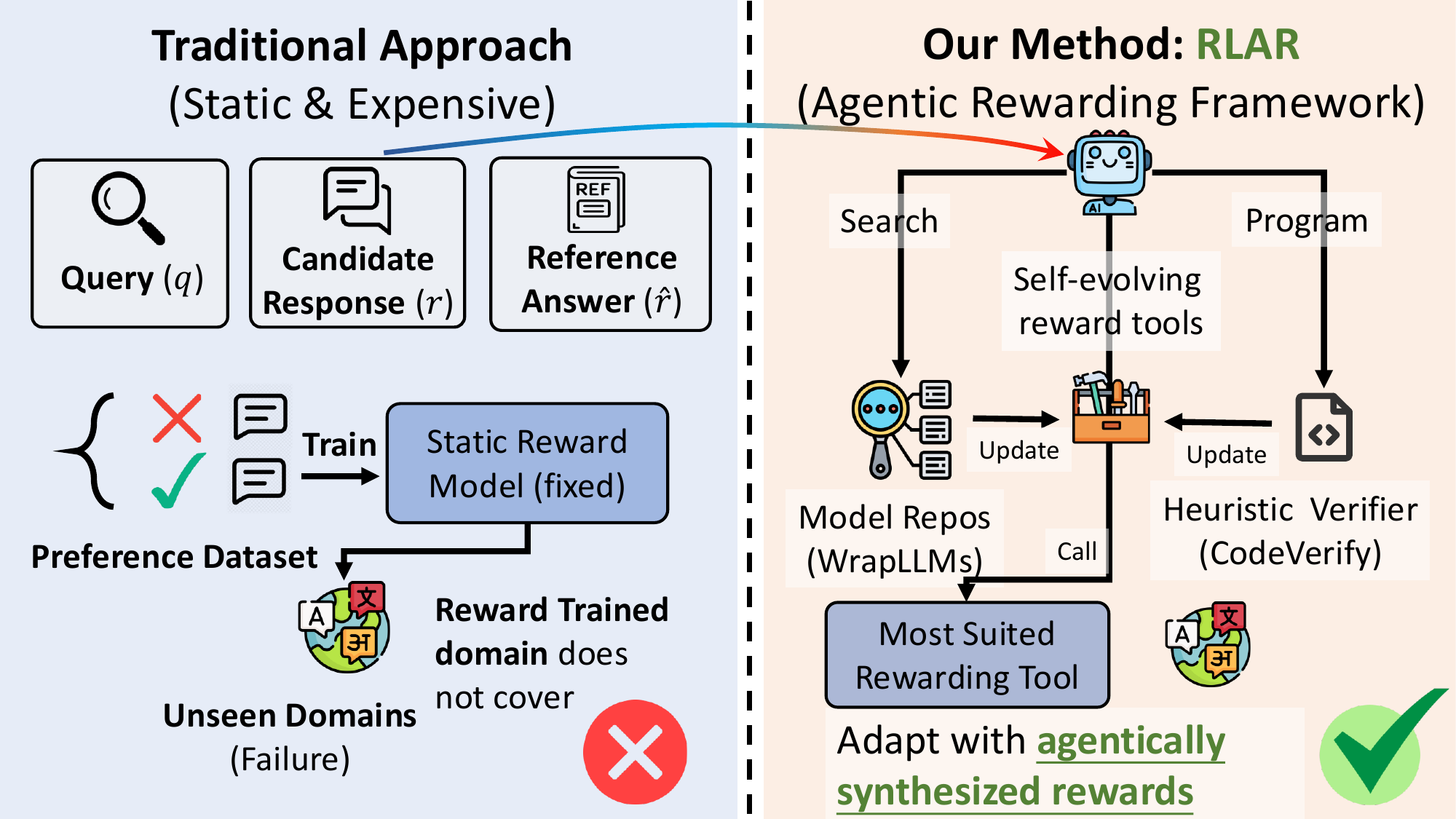}
    \caption{A conceptual comparison between traditional reward modeling and our proposed RLAR framework.}
    \label{fig:intro_fig}
    \vspace{-4mm}
\end{figure}

Currently, trending developing practices of building a reward system are to create a static judge model~\citep{skyworkreward, tulu3}, including gathering a wide range of training data, and the judge would therefore cover the distribution. 
However, this practice faces the out-of-distribution challenges when shifting reward models to unseen domains~\citep{toolrm, nofreelabels}. 
Meanwhile, gathering reward training data is hindered by prohibitive annotation costs, the need for niche domain expertise~\citep{lessons-mathprm, li-etal-2025-codeprm}, and susceptibility to introducing bias from distillation of synthetic data~\citep{llmarenotfairevaluators, preferenceleakage, nofreelabels}.

We aim at bridging the gap for a high-quality and generalizable rewarding system at a controlled cost without much expertise or effort. We propose \textbf{\method{}} (Reinforcement Learning from Agentic Rewards), an automated framework capable of dynamically scaling the reward system as the training data extends. 
We leverage the tool-calling capabilities of LLMs to enable the agentic exploration of high-quality reward model resources across the Internet model repositories. These specialized models typically outperform generic reward models on their intended tasks, yet they remain an underutilized resource. Additionally, we incorporate code-generation abilities to synthesize automated verifiers for math and coding-related tasks, eliminating the possible loopholes introduced by large language reward models.
By shifting from learning from LLMs' ``\textbf{score-form feedback}'' to its ``\textbf{designed reward functions}'', \method{} ensures high-fidelity reward signals with domain shifts while significantly reducing the inference costs associated with construction of additional reward models.

We evaluate the \method{} across a diverse suite of tasks, including math-reasoning~\citep{hendrycks2021measuringmathematicalproblemsolving, gsm8k}, code~\citep{leetcodedataset, mbpp}, translation~\citep{benchmax, nllb2022, flores101, twoevaldatasets} and general chat~\citep{ultrachat}. We compare our agentic reward synthesis-leveraging framework against state-of-the-art static reward models, including classification \citep{skyworkreward}, generative ones \citep{skyworkcritic2024} and GPT5-as-a-judge. 
\method{} achieves overall performance improvements of \(10.4\%\) and \(61.9\%\) on Llama-3.1-8B and Qwen3-8B, significantly surpassing static classification and generative baselines. Unlike baselines such as \textbf{SkyLlama}, which suffer from performance decrements in math reasoning tasks during mixed training, \method{} dynamically models domain objectives across math, code, translation, and chat. 
Notably, \method{} demonstrates enhanced robustness to \textbf{format and verbosity hacking}, while there is still loopholes in baselines, even the GPT5-as-a-judge.
Finally, compared to GPT-5-as-a-judge, \method{} reduces API token consumption by approximately \(80\%\) and  GPU training hours by \(75\%\), demonstrating \method{}'s scaling potentials.

Further experiments demonstrate that the effectiveness of RLAR mainly comes from the accurate reward tool selection, and we calibrate it with \textsc{RewardBench-v2}. Our \textbf{Agentic Reward Tool Selector} obtained \(90.44\%\) accuracy by routing queries to the most suitable reward models, outperforming both the SOTA \(87.19\%\) and SOTA-model logits ensembling \(87.44\%\). These results indicate that dynamic routing more effectively approaches the theoretical upper bound.

\section{Preliminaries}

\subsection{LLM post-training using RL}

In reinforcement learning from human feedback (RLHF), the \textbf{Proximal Policy Optimization} algorithm~\citep{ppo} is frequently employed for policy optimization.  
The typical workflow begins with a \textit{warm-start training} phase in which a \textbf{Value Model} (often a reward model) is learned. Training triplets \((x, y_{+}, y_{-})\) are sampled from human preference data, with \(y_{+}\) being the preferred over \(y_-\). Let \(r_{\theta}(x, y)\) be the reward assigned by model given prompt \(x\) and response \(y\), the learning objective function can be expressed as:
\[
\mathcal{L}(\theta) = -\frac{1}{N} \,\mathbb{E}_{(x,y_+, y_-) \sim D } \bigg[ \log \sigma \big( \Delta r_{\theta} \big) \bigg],
\]
where \(\sigma(\cdot)\) is the logistic sigmoid function and \(\Delta r_{\theta} = r_{\theta}(x, y_+) - r_{\theta}(x, y_-)\)  .

We research a more training-efficient framework. The \textbf{Group Relative Policy Optimization} (GRPO)~\citep{grpo} modifies the advantage estimation to reduce the dependence on a learnable value model for estimating the advantage baseline. Instead, GRPO computes the normalized advantage within a group of sampled outputs:
\[
\hat{A}_{i} = \frac{r_i - \mathrm{mean}(r)}{\mathrm{std}(r)},
\]
where \(r=\{ r_i \}_{i=1}^G\) are the rewards assigned to \(G\) candidate outputs for the same prompt, \(\mathrm{mean}(r)\) and \(\mathrm{std}(r)\) are computed over the group. Also, the Kullback-Leibler Divergence term is removed from the per-step reward and is instead applied directly to the overall optimization objective:

\begin{table*}[t]
    \centering
    \small 
    \begin{tabular}{llccccl}
    \toprule
    \textbf{Domain} & \textbf{Dataset} & \textbf{Train Size} & \textbf{Test Size} & \textbf{OOD} & \textbf{Evaluation Metric} \\
    \midrule
    \multirow{3}{*}{\textbf{Math}} 
        & \textsc{GSM8k}          & 1,500 & 1,319 & \xmark & \multirow{3}{*}{Numeric Match} \\
        & \textsc{Hendrycks-Math} & 1,500 & 500   & \xmark &  \\
        & \textsc{AIME-24/25}        & 0     & 60    & \cmark &  \\
    \midrule
    \multirow{2}{*}{\textbf{Code}} 
        & \textsc{LeetCode-Dataset}       & 2,585 & 223   & \xmark & \multirow{2}{*}{Unit Test Execution} \\
        & \textsc{MBPP}           & 0     & 974   & \cmark &  \\
    \midrule
    \multirow{2}{*}{\textbf{Translation}} 
        & \textsc{Flores-200}     & 3,000 & 300   & \xmark & \multirow{2}{*}{Hybrid (LLM-Judge + BLEU)} \\
        & \textsc{WMT-24}         & 0     & 500   & \cmark &  \\
    \midrule
    \textbf{General} 
        & \textsc{UltraChat}      & 2,500 & 200   & \xmark & Model-based (LLM-as-a-Judge) \\
    \bottomrule
    \end{tabular}
    \caption{Summary of domains and datasets considered across \textbf{Math}, \textbf{Code}, \textbf{Translation}, and \textbf{General} Conversation.}
    \label{tab:dataset_stat}
\end{table*}

\begin{align*}
\max_{\phi} \; \;
& \mathbb{E}_{\substack{x \sim D, \\ \{y_i\}_{i=1}^G \sim \pi_{\mathrm{old}}(y_i|x)}} 
\Bigg[
   \frac{1}{G} \sum_{i=1}^G
   \min \Bigg\{
      \frac{\pi_{\theta}(y_i|x)}{\pi_{\mathrm{old}}(y_i|x)} ,  \;
      \nonumber \\[-0.5em]
& 
      \mathrm{clip}\!\left(
         \frac{\pi_{\theta}(y_i|x)}{\pi_{\mathrm{old}}(y_i|x)},
         1 - \epsilon, \;
         1 + \epsilon
      \right) 
   \Bigg\} \hat{A}_{i},
\Bigg] \nonumber \\[0.5em]
& \quad - \beta \, \mathbb{D}_{\mathrm{KL}}\!\left[ \pi_{\phi} \;\|\; \pi_{\mathrm{ref}} \right].
\end{align*}
This formulation reduces sensitivity to reward model estimation errors by leveraging relative comparisons within output groups.

\subsection{Task Design and Evaluation} \label{sec:task_design}

We studied the blended domains where real LLM post-training usually focuses on: \textbf{mathematical reasoning}, \textbf{code and programming}, \textbf{translation}, and \textbf{general conversation} tasks. We aim to promote the importance of cross-domain generalization and expose the need to dynamically design a customized reward function for diverse task domains. We show information about the datasets used in our experiment in Table~\ref{tab:dataset_stat}.

\section{Methodology}

\begin{figure*}[t]
    \centering
    \includegraphics[width=0.8\linewidth]{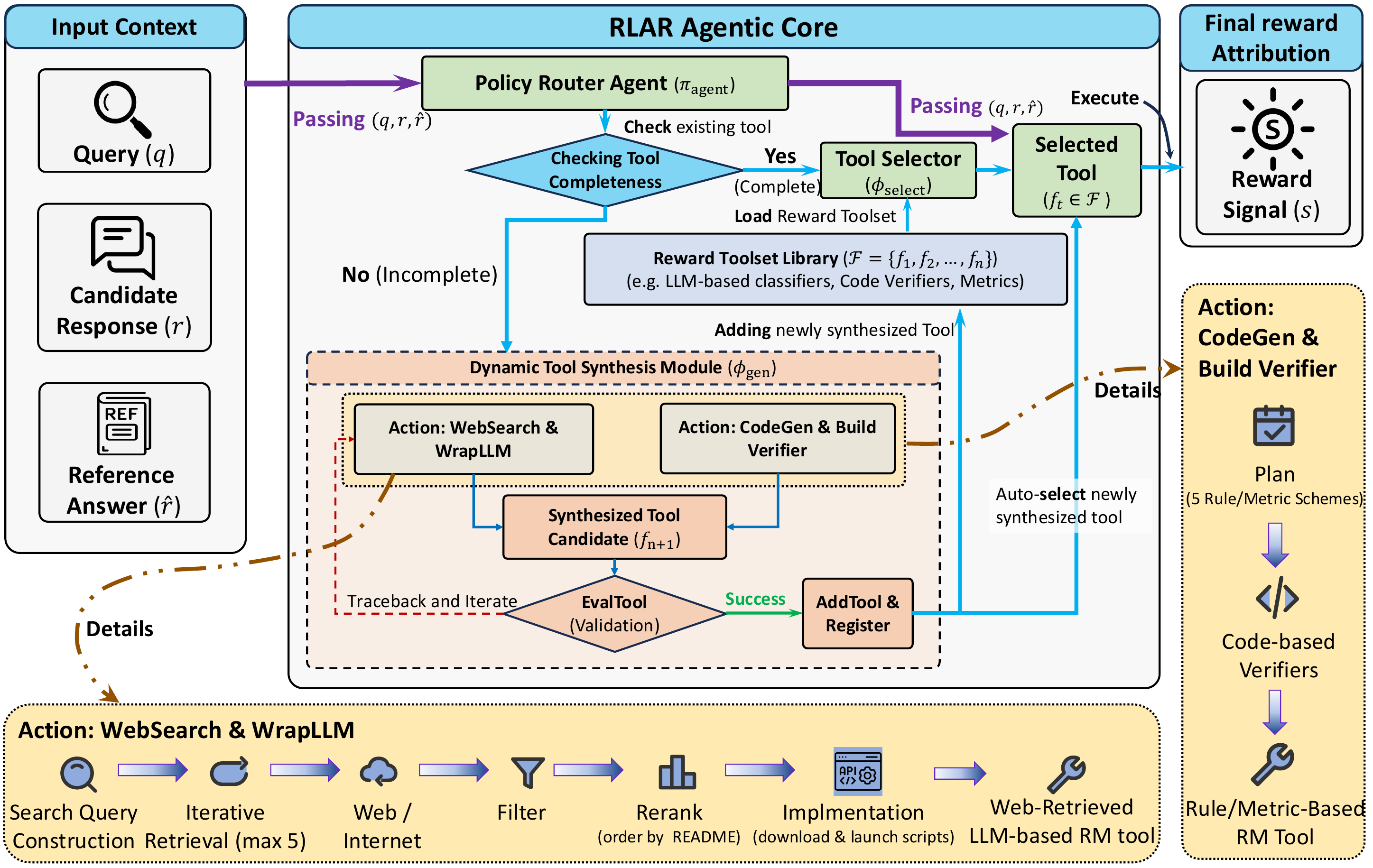}
    \caption{The Reinforce Learning with Agentic Reward (RLAR) Framework.}
    \label{fig:main_fig}
    \vspace{-4mm}
\end{figure*}

To address the inherent limitations of monolithic reward models in handling heterogeneous task distributions, we propose \textbf{Reinforcement Learning with Agentic Reward} (\method{}). Unlike static reward functions, \method{} is a self-evolving framework that leverages the tool-use capabilities of LLMs to dynamically synthesize, verify, and invoke specialized reward tools.

\subsection{Problem Formulation} \label{sec:formulation}
We formalize the \method{} framework as a dynamic reward synthesis and calling problem. Given an input context triplet $(q, r, \hat{r})$ comprising a query, the candidate response, and the reference answer, our objective is to compute a precise scalar reward $s \in \mathbb{R}$ that reflects the response quality.

The framework operates on a \textbf{Reward Toolset (Library)} $\mathcal{F} = \{f_1, f_2, \dots, f_n\}$, where each $f_i: (q, r, \hat{r})\rightarrow s$ is an atomic reward tool  (e.g., an LLM-classifier-based Reward Model, a code-based checker, or a heuristic rule). Initially, $\mathcal{F}$ is populated with a general-purpose base reward model (e.g., Skywork-Llama). As the framework encounters more queries from diverse tasks, $\mathcal{F}$ expands through agentic synthesis: $\mathcal{F}_{t+1} = \mathcal{F}_t \cup \{f_{new}\}$.

\subsection{Adaptive Reward Routing and Selection} \label{sec:routing}
A core challenge in \method{} is determining whether the current toolset $\mathcal{F}$ can accurately evaluate a given \((q, r, \hat{r})\). We introduce a \textbf{Policy Router} $\pi_{agent}$, which serves as the decision-making core.

\noindent\textbf{Completeness Assessment:} For each incoming $(q, r, \hat{r})$, $\pi_{agent}$ assesses the \textit{competency} of the existing library. We formalize this as a classification-of-intent task:
\begin{equation}
    a = \pi_{agent}(\text{state} = (q, r, \hat{r}), \mathcal{F})
\end{equation}
where $a \in \{\text{Select}, \text{Synthesize}\}$.

\noindent\textbf{Tool Selection:} If $\mathcal{F}$ is complete ($a = \text{Select}$), a selection module $\phi_{select}$ identifies the optimal reward tool $f_t \in \mathcal{F}$ by matching the task metadata with tool descriptions.

\noindent\textbf{Tool Synthesis:} If no suitable tool exists ($a = \text{Synthesize}$), it calls the Agentic Synthesis module to construct the missing reward tool.

\subsection{Agentic Tool Synthesis} \label{sec:synthesis}

When a tool is missing from the reward tool library \(\mathcal{F}\), \method{} invokes specialized agents to construct a new reward tool $f_{new}$. Considering different reward system types, we designed \textbf{WrapLLM} and \textbf{CodeVerify} to fulfill this goal. 

\subsubsection{WrapLLM}

The LLM agent retrieves and encapsulates the most compatible lightweight (within 10B) Reward Model repository from the open-sourced model platform \textit{HuggingFace}. They suit tasks requiring semantic nuances, creativity, or value alignment. The agent follows an iterative retrieval workflow: 
(1) Generate a search query for the search engine
(2) Iteratively retrieving from the search engine
(3) Filter and rerank results for the most related reward model
(4) Download and implement the deployment and inference code 
(5) Wrap (generating the function names, descriptions) into an API for tool calling. 
The workflow transforms a reward model checkpoint into a callable tool for $\mathcal{F}$. 

\subsubsection{CodeVerify}

The agent follows a Plan-and-Execute strategy to generate deterministic verifiers for objective tasks such as mathematics and programming. It would initially plan at most 5 possible schemes and implement the optimal one into reward tool. 

The agent generates Python-based scripts that vary from task type. For example, retrieving the final answer from a formatted solution as a reward tool is suited for math; retrieving and arranging generated code and executing it with test cases would be suited for code. Furthermore, for tasks such as translation, it would implement a metric (e.g., ROUGE, BLEU, METEOR) based reward tool.  The prompts for the above are provided in Appendix~\ref{apd:prompt_wrap_code}.


\subsection{Verification and Library Update} \label{sec:verification}

To ensure the reliability of the synthesized tools, each $f_{new}$ must pass a module (\textbf{EvalTool}) before being committed to $\mathcal{F}$. 
For \textit{WrapLLM}, it examines the consistency between the model-as-a-tool description and its original documentation. For \textit{CodeVerify}, it performs analysis on code implementation and execution checks on synthesized scripts. Only tools that satisfy these verifications are committed to the library: $\mathcal{F} \leftarrow \mathcal{F} \cup \{f_{new}\}$.

\section{Experiments}

\subsection{Dataset and Evaluation}

\noindent\textbf{Math Reasoning:} We use \textsc{GSM8k}~\citep{gsm8k} and \textsc{Hendrycks-MATH}~\citep{hendrycks2021measuringmathematicalproblemsolving} for mathematical reasoning. The goal is to reason for the final numeric solution. 

\noindent\textbf{Coding:} We select the \textsc{LeetCode-Dataset}~\citep{leetcodedataset}, which asks the testers to fill in the LeetCode cloze-style function implementation.

\noindent\textbf{Translation:} We sample from \textsc{FLORES-101}~\citep{flores101} to cover 20 translation pairs across five languages (CN, DE, EN, FR, JP, in alphabetical order). This task requires mastery across languages and is important for LLMs' accessibility to worldwide users, especially non-English speakers.

\noindent\textbf{General Alignment:} \textsc{UltraChat}~\citep{ultrachat} is employed to train and evaluate multi-turn dialogue and instruction-following capabilities.

\noindent\textbf{Generalization Challenge.}\quad To assess the robustness of the trained policy, we applied benchmarks including \textsc{AIME-24/25} for math reasoning, \textsc{MBPP}~\citep{mbpp} for Python synthesis, and \textsc{WMT-24} for unseen translation contexts. 
These benchmarks evaluate that the post-trained model can generalize rather than over-fit to the specific heuristics of the training reward models, as an indication of reward hacking.

\noindent\textbf{Data Process.} \quad For all queries in the training dataset split,  we applied an automated quality filtering process, requiring the LLM to remove samples that did not meet the standards based on both query quality and response quality. The prompt is provided in Appendix~\ref{apd:prompt_for_data_filter}. Then we uniformly downsample over \textsc{GSM8k, Hendrycks-Math, Flores-200, WMT-24, Ultra-Chat} to ensure a balanced distribution of queries across different dataset sources. To facilitate our experiments, we obtained the reasoning outputs of \textsc{GPT-5} on all queries. For dataset queries lacking human-annotated reasoning references, we supplemented them with the results from \textsc{GPT-5}. Table~\ref{tab:dataset_stat} shows the statistics of the train and validation sets. A more detailed introduction of the task and our selected dataset is listed in Appendix~\ref{apd:data}.

\noindent\textbf{Evaluation Metrics.} \quad To ensure evaluation rigor, we apply Numeric Exact Match (\textbf{NEM}) at Best@1 for mathematical reasoning tasks.  For code generation, we report \textbf{pass@1} verified by automated code executors.
For translation tasks, we report both \textbf{BLEU}~\citep{papineni-etal-2002-bleu} and \textbf{LLM-judge} scores, while general dialogue tasks are evaluated solely by the LLM-judge. Additionally, we measure the \textbf{average generation lengths} across all benchmarks to detect potential reward-hacking behaviors associated with the verbosity bias inherent in LLM-based rewarding. GPT-5 is utilized as the primary evaluator for all LLM-as-a-judge assessments, and the 0-10 scoring range is normalized to 0-100 in accordance with other metrics.
We provide an \textbf{average score} computed by averaging individual benchmark results. Specifically, for the translation benchmark, the score is a composite of 0.5 BLEU-2 and 0.5 LLM-judge.

\begin{table*}[t]
    \centering
    \resizebox{\textwidth}{!}{%
    \begin{tabular}{l c ccccc cccc ccc}
    \toprule
    \textbf{Domain} & \textbf{Avg.} & \multicolumn{4}{c}{\textbf{Math}} & \multicolumn{2}{c}{\textbf{Code}} & \multicolumn{4}{c}{\textbf{Translation}} & \multicolumn{2}{c}{\textbf{General}} \\
    \cmidrule(r){1-1} \cmidrule(r){3-6} \cmidrule(r){7-8} \cmidrule(r){9-12} \cmidrule(r){13-14}
    \textbf{Dataset} & \textbf{Score} & GSM8k & HEND & \textbf{AIME24} & \textbf{AIME25} & LCode & \textbf{MBPP} & \multicolumn{2}{c}{Flore} & \multicolumn{2}{c}{\textbf{WMT24}} & UChat & Avg Len \\ \midrule
    \textbf{Metric} & - & NEM & Pass@k & NEM & NEM & Pass@k & Pass@k & BLEU-2 & LLM-J & BLEU-2 & LLM-J & LLM-J & \# \\
    \midrule
    \multicolumn{14}{c}{\cellcolor{lightgreen}{\textit{\textbf{Llama-3.1-8B}}}} \\ \midrule
    Base & 37.81 & 60.50 & 49.80 & 6.67 & 0.00 & \underline{11.21} & 41.17 & 21.04 & 77.27 & 26.39 & 79.24 & \underline{68.97} & 152.84 \\
    SkyLlama & 23.47 & 11.30 \down{49.20} & 47.60 \down{2.20} & 3.33 \down{3.34} & 0.00 \neut{-} & 8.97 \down{2.24} & 7.80 \down{33.37} & 21.16 \up{0.12} & 45.32 \down{31.95} & 23.23 \down{3.16} & 54.68 \down{24.56} & 60.03 \down{8.94} & 147.39 \\
    SeedX & 17.29 & 1.21 \down{59.29} & 42.20 \down{7.60} & 6.67 \neut{-} & 0.00 \neut{-} & 0.00 \down{11.21} & 13.76 \down{27.41} & 0.00 \down{21.04} & 23.24 \down{54.03} & 0.00 \down{26.39} & 42.76 \down{36.48} & 58.73 \down{10.24} & 89.95 \\ 
    SkyQwen & 31.68 & 16.15 \down{44.35} & 46.60 \down{3.20} & 3.33 \down{3.34} & 0.00 \neut{-} & \textbf{11.66 \up{0.45}} & 36.04 \down{5.13} & 26.42 \up{5.38} & 77.53 \up{0.26} & \underline{29.80 \up{3.41}} & 78.74 \down{0.50} & 65.13 \down{3.84} & 134.21 \\ 
    SkyCritic & 34.05 & 44.58 \down{15.92} & 48.00 \down{1.80} & \underline{10.00 \up{3.33}} & 0.00 \neut{-} & 10.31 \down{0.90} & \textbf{45.69 \up{4.52}} & 25.31 \up{4.27} & 42.53 \down{34.74} & 28.21 \up{1.82} & 67.82 \down{11.42} & 65.93 \down{3.04} & 106.97 \\ 
    GPT5-Judge & \underline{39.89} & \underline{71.95 \up{11.45}} & 43.20 \down{6.60} & 6.67 \neut{-} & \underline{6.67 \up{6.67}} & 7.17 \down{4.04} & \underline{45.17 \up{4.00}} & 20.35 \down{0.69} & \textbf{82.50 \up{5.23}} & 26.37 \down{0.02} & \textbf{84.42 \up{5.18}} & \textbf{71.32 \up{2.35}} & 203.31 \\ 
    Human & 33.56 & 42.38 \down{18.12} & 40.00 \down{9.80} & 3.33 \down{3.34} & 3.33 \up{3.33} & 10.31 \down{0.90} & 37.37 \down{3.80} & \underline{26.61 \up{5.57}} & 70.67 \down{6.60} & \textbf{29.81 \up{3.42}} & 75.88 \down{3.36} & 63.87 \down{5.10} & 76.76 \\
    \textbf{RLAR}\small{(Ours)} & \textbf{41.73} & \textbf{73.61 \up{13.11}} & \textbf{54.60 \up{4.80}} & \textbf{13.33 \up{6.66}} & \textbf{6.67 \up{6.67}} & \textbf{11.66 \up{0.45}} & 43.18 \up{2.01} & \textbf{26.73 \up{5.69}} & \underline{74.83 \down{2.44}} & \underline{29.80 \up{3.41}} & \underline{79.54 \up{0.30}} & 67.03 \down{1.94} & 141.85 \\
    \midrule
    \multicolumn{14}{c}{\cellcolor{lightgreen}{\textit{\textbf{Qwen3-8B}}}} \\ \midrule
    Base & 29.12 & 31.99 & 19.80 & 3.33 & 0.00 & 6.28 & 14.27 & 28.46 & 84.00 & 32.50 & 84.46 & 71.70 & 1064.04 \\
    SkyLlama & 29.25 & 0.00 \down{31.99} & 35.80 \up{16.00} & 10.00 \up{6.67} & 6.67 \up{6.67} & 7.62 \up{1.34} & 14.85 \up{0.58} & 28.78 \up{0.32} & 85.59 \up{1.59} & 33.75 \up{1.25} & 86.54 \up{2.08} & 71.00 \down{0.70} & 844.96 \\
    SeedX & 35.14 & 0.00 \down{31.99} & \textbf{72.60 \up{52.80}} & 6.67 \up{3.34} & 6.67 \up{6.67} & 21.08 \up{14.80} & 20.53 \up{6.26} & 30.34 \up{1.88} & 84.97 \up{0.97} & 35.25 \up{2.75} & 85.14 \up{0.68} & 70.83 \down{0.87} & 1135.82 \\
    SkyQwen & 33.48 & 0.00 \down{31.99} & 67.00 \up{47.20} & 3.33 \neut{-} & 3.33 \up{3.33} & 15.70 \up{9.42} & \underline{21.25 \up{6.98}} & 29.70 \up{1.24} & 85.72 \up{1.72} & 33.82 \up{1.32} & 85.18 \up{0.72} & 73.47 \up{1.77} & 1192.12 \\ 
    SkyCritic & 39.07 & 51.71 \up{19.72} & 58.00 \up{38.20} & \textbf{20.00 \up{16.67}} & 6.67 \up{6.67} & 13.00 \up{6.72} & 12.32 \down{1.95} & 28.66 \up{0.20} & 86.09 \up{2.09} & 32.52 \up{0.02} & 86.46 \up{2.00} & 73.10 \up{1.40} & 691.97 \\
    GPT5-Judge & \underline{45.85} & \underline{91.96 \up{59.97}} & 56.20 \up{36.40} & 16.67 \up{13.34} & \underline{10.00 \up{10.00}} & \underline{21.52 \up{15.24}} & 18.17 \up{3.90} & 28.31 \down{0.15} & \textbf{89.77 \up{5.77}} & 32.64 \up{0.14} & \textbf{88.93 \up{4.47}} & \textbf{78.32 \up{6.62}} & 1462.29 \\
    Human & 44.40 & \textbf{95.30 \up{63.31}} & 53.80 \up{34.00} & \textbf{20.00 \up{16.67}} & \underline{10.00 \up{10.00}} & 8.52 \up{2.24} & 18.89 \up{4.62} & \underline{30.71 \up{2.25}} & 85.11 \up{1.11} & \underline{35.91 \up{3.41}} & 85.22 \up{0.76} & 74.63 \up{2.93} & 692.71 \\ 
    \textbf{RLAR}\small{(Ours)}  & \textbf{47.16} & 93.70 \up{61.71} & \underline{59.80 \up{40.00}} & \textbf{20.00 \up{16.67}} & \textbf{13.33 \up{13.33}} & \textbf{23.31 \up{17.03}} & 19.30 \up{5.03} & \textbf{31.12 \up{2.66}} & \underline{86.49 \up{2.49}} & \textbf{36.07 \up{3.57}} & \underline{86.67 \up{2.21}} & \underline{74.81 \up{3.11}} & 986.23 \\
    \bottomrule
    \end{tabular}}
    \caption{\textbf{Main Experiment Results}. For each testset, \textbf{Bold} indicates the best result under that metric, and \underline{underline} indicates the second best within each model group. \up{Increment} and \down{decrement} are calculated relative to the Zero-Shot baseline for Llama and Qwen. \textbf{LLM-J} is short for LLM-judge, and \textbf{NEM} shorts for numeric exact match. \textbf{HEND} shorts for \textsc{Hendrycks-Math}, \textbf{LCode} shorts for \textsc{LeetCode-Dataset}, \textbf{UChat} shorts for \textsc{Ultra-Chat}. The bold benchmarks (\textbf{AIME24/25, MBPP, WMT24}) indicate that they do not include the training data.}
    \label{tab:main_res}
\end{table*}

\subsection{Experimental Setup} \label{sec:experimental_setup}

\textbf{Base Models}: We use Qwen3-8B \citep{qwen3report} and Llama-3.1-8B-Instruct \citep{llama3report} as base models. To examine various reward system architectures, we incorporate the following baselines:

\noindent\textbf{Classification Reward Baselines:} For the single classification reward model setting, we select \texttt{Skywork-Reward-V2-Llama-3.1-8B} (\textbf{SkyLlama}) and \texttt{Skywork-Reward-V2-Qwen3-8B} (\textbf{SkyQwen}) \citep{skyworkreward}, both of which represent the state-of-the-art on \textsc{RewardBench-v1/v2}~\cite{rewardbench, rewardbench2}. Additionally, we incorporate \texttt{Seed-X-8B-RM} (\textbf{SeedX}) due to its robust performance in multilingual contexts~\citep{seedX}. 

\noindent\textbf{Generative Reward Baselines}: We evaluate the \texttt{Skywork/Skywork-Critic-Llama-3.1-8B} (\textbf{SkyCritic}) \citep{skyworkcritic2024} , which generates natural language justifications with its judgments. Furthermore, we employ GPT-5-as-a-judge (\textbf{GPT5-judge}) as a high-capacity competitor, following the RL from AI Feedback paradigm~\citep{lee2024rlaif}. In this setup, GPT-5 is prompted to act as an evaluator, and its numerical scores are extracted as reward signals. The specific prompt template is detailed in Appendix~\ref{apd:judge_prompt}.

\noindent\textbf{Human Baseline:}
We implement a task-specific reward system to calibrate performance against optimal human-defined heuristics. For each task, the reward function is defined by the metrics most appropriate for the domain. Specifically, we utilize \textbf{NEM} for mathematics:
\[
    R_{\text{math}}(r, \hat{a}) =  \begin{cases}  1, & \text{if } \text{NEM}(r, \hat{a}) \\  0, & \text{otherwise} \end{cases}
\]
where $\hat{a}$ denotes the ground-truth numerical answer,
and \textbf{pass@1} for coding:
\[
R_{\text{code}}(r) =  \begin{cases}  1, & \text{if } \text{Exec}(r) = \text{Pass} \\  0, & \text{otherwise} \end{cases}
\]
For translation, the reward is calculated as a hybrid score \(w_1 \cdot \mathrm{BLEU} + w_2 \cdot \mathrm{SeedX} \), where the scaling factor is $w1 = 1/2, w_2 = 1/2$ and SeedX's score is normalized to \((0, 1)\). The dialogue tasks are rewarded only by \textbf{SkyLlama}. We have attached the \textbf{training configurations} in Appendix~\ref{sec:training_config} and \textbf{hyperparameter} for training in Appendix~\ref{apd:training_details}. \method{} is powered by GPT-5 by default.

\subsection{Main Results}

\noindent\textbf{Superiority of RLAR across Mixed Domains.} \quad
RLAR consistently outperforms both classification-based and generative reward baselines across multiple tasks and backbone models (average score \(+10.4\%\) on Llama and \(+61.9\%\) on Qwen). 
Notably, RLAR achieves the best results in the majority of benchmarks ($73.6$ NEM on GSM8k for Llama-3.1 and $23.3$ Pass@1 on LeetCode for Qwen3) without incurring significant performance degradation in other data distributions.
The increase can also be found in the OOD benchmarks \textsc{AIME-24}(\(+16.6\)),  \textsc{AIME-25}(\(+13.3\)), \textsc{AIME-24}(\(+5.03\)) and \textsc{WMT-24}(\(+3.41\) BLEU).
In contrast, \textbf{SkyLlama}, \textbf{SkyQwen}, \textbf{SeedX} suffer from catastrophic failures when faced with Qwen3 on \textsc{GSM-8k} (\(-31.9\)) and \textsc{UltraChat} while they exhibits benefits to \textsc{Hendyrcks} and \textsc{LeetCode}. This indicates that \method{} maintains stable and generalizable performance by dynamically modeling domain objectives into new reward function designs without neglecting specific distributions.

The results from Llama-3.1 indicate that RL post-training might not be beneficial, probably due to its near-saturated base model performance ($60.5$ on GSM8k, $41.17$ on MBPP). However, \method{} still causes most increases on most in-distribution and OOD test-benches ($+13.1$ on GSM-8k, $+3.41$ on \textsc{WMT} evaluated by BLEU). \textbf{SkwLlama}, \textbf{SkyQwen}, \textbf{SkyCritic} mostly failed on GSM8k, causing catastrophic performance loss (\(-49.2, -44.3, -15.9\)). 
They also rarely show benefits on OOD benchmarks (AIME, MBPP, WMT), indicating the performance gain is less often generalized outside the training domains.
\textbf{GPT5-Judge} optimized LLama on \textsc{GSM8k} by \(11.4\), while causing degrades on \textsc{Hendrycks-Math} by \(6.60\) and \textsc{LeetCode} by \(4.04\). These demonstrate \method{}'s robustness to avoid performance degradation, benefiting from its more comprehensive reward design.

\noindent\textbf{Comparison with GPT5-Judge and Human Baseline.} \quad
While\textbf{ GPT5-judge} serves as a competitive baseline, it exhibits vulnerabilities in rule-heavy tasks. For example, on Llama-3.1, GPT5-Judge triggers performance drops on Hendrycks-Math and LeetCode compared to zero-shot levels (\(-6.60\)/\(-4.04\)) while RLAR does not (\(+4.80\)/\(+0.45\)), suggesting that reward systems lacking explicit rule verification are susceptible to ``blind spots'' in deterministic domains. Furthermore, RLAR demonstrates significant advantages in \textbf{cost-effectiveness and training efficiency}: a full 80-step RL training session with GPT-5 costs approximately 120 million API tokens, and the training takes over GPU 288 hours, whereas RLAR reduces them to 23.7 million API tokens and 72 GPU hours. RLAR represents a highly efficient and economical paradigm for RLAIF.


Intriguingly, RLAR even surpasses the Human baseline in several reasoning tasks (Math and Code benchmarks on Llama, \textsc{Hendrycks-Math, LeetCode} on Qwen). Qualitative analysis reveals that RLAR's reward signals provide more comprehensive coverage than simple heuristics, incorporating additional test cases and rigorous mathematical formatting checks. This effectively increases the completeness of the rule-based verifiers.

\subsection{Reward Hacking Identification}

\noindent \textbf{Hack with Format.} \quad
A critical failure mode observed in classification-based RM baselines on Qwen3-8B is ``Format Hacking.'' Specifically, all three classification baselines (\textbf{SkyLlama, SeedX, SkyQwen}) collapsed to a \(0.00\) score on \textsc{GSM8k}. Manual inspection reveals that the policy models learned to bypass task-specific constraints: while the instructions required ``\#\#\#\#'' before \textsc{GSM8k} answers and ``\texttt{\textbackslash{}boxed\{\}}'' for \textsc{Hendrycks-Math}, the models mistakenly unified all outputs into the Hendrycks format. 
Thus, baselines may overlook subtle instructional requirements, allowing the policy to 'exploit' the reward system by mimicking specific output formats rather than optimizing the actual reward.
Our method exhibits robustness on such loopholes, as the trained policy can easily handle the formats across all benchmarks.

\noindent\textbf{Hack with Verbosity.} \quad
Furthermore, we observe a persistent verbosity bias in LLM-based rewards baselines. For example, models trained with GPT-5-Judge consistently produced outputs \(30\%\) \textbf{longer} than those partly learned from rule-based rewards (Human/RLAR). This confirms that verbosity bias is inherently latent in reward LLMs and is potentially triggered during RL. 
Due to the rewarding versatility and the abundant included verifiable rewards, \method{} did not introduce a significant long-tail generation problem.

\subsection{Ablations on Modules}

\begin{figure}[ht]
     \centering
     \begin{subfigure}{0.22\textwidth}
         \centering
         \includegraphics[width=\linewidth]{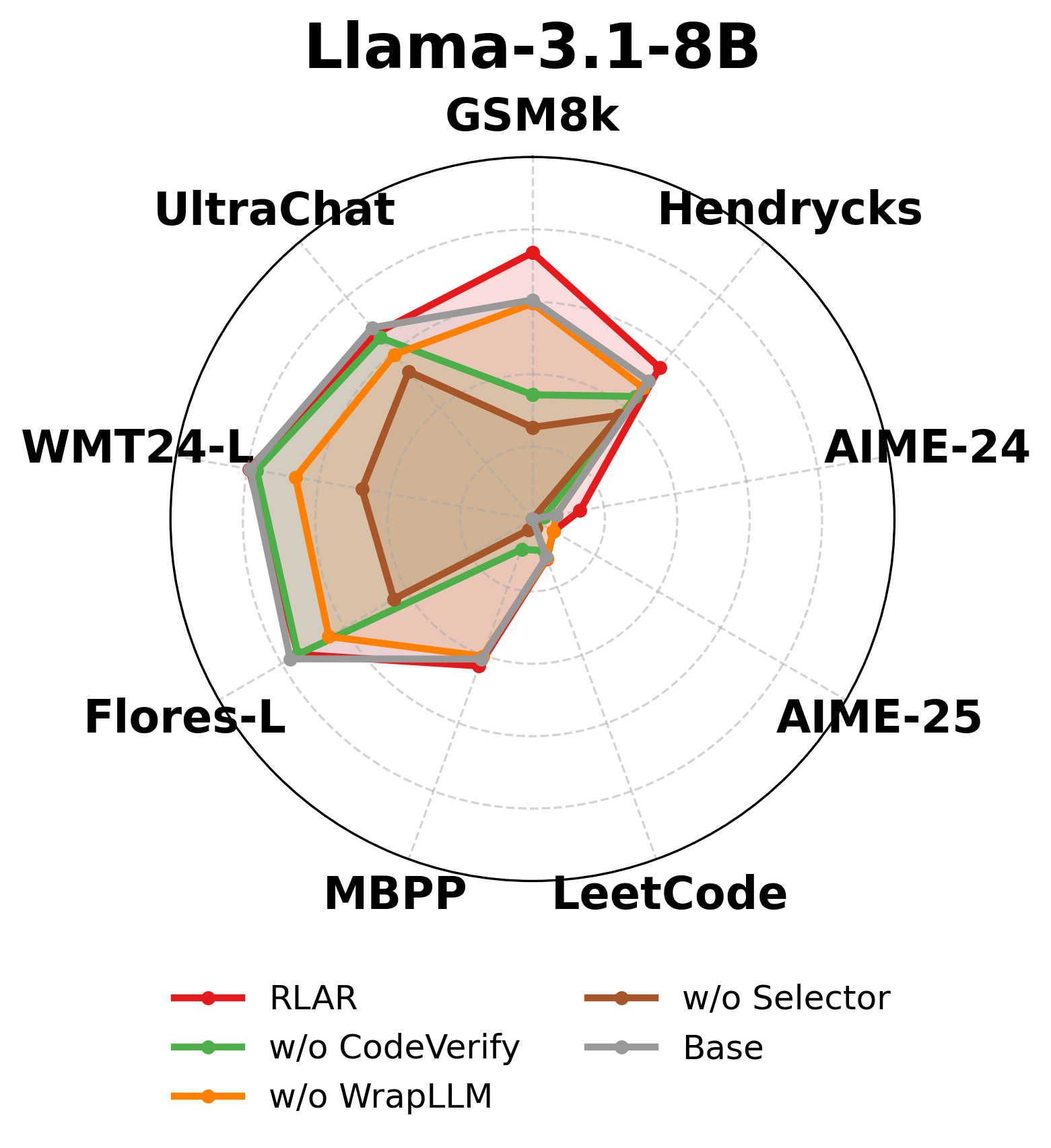}
         \caption{Llama-3.1-8B}
         \label{fig:abl_llama}
     \end{subfigure}
     \hfill
     \begin{subfigure}{0.22\textwidth}
         \centering
         \includegraphics[width=\linewidth]{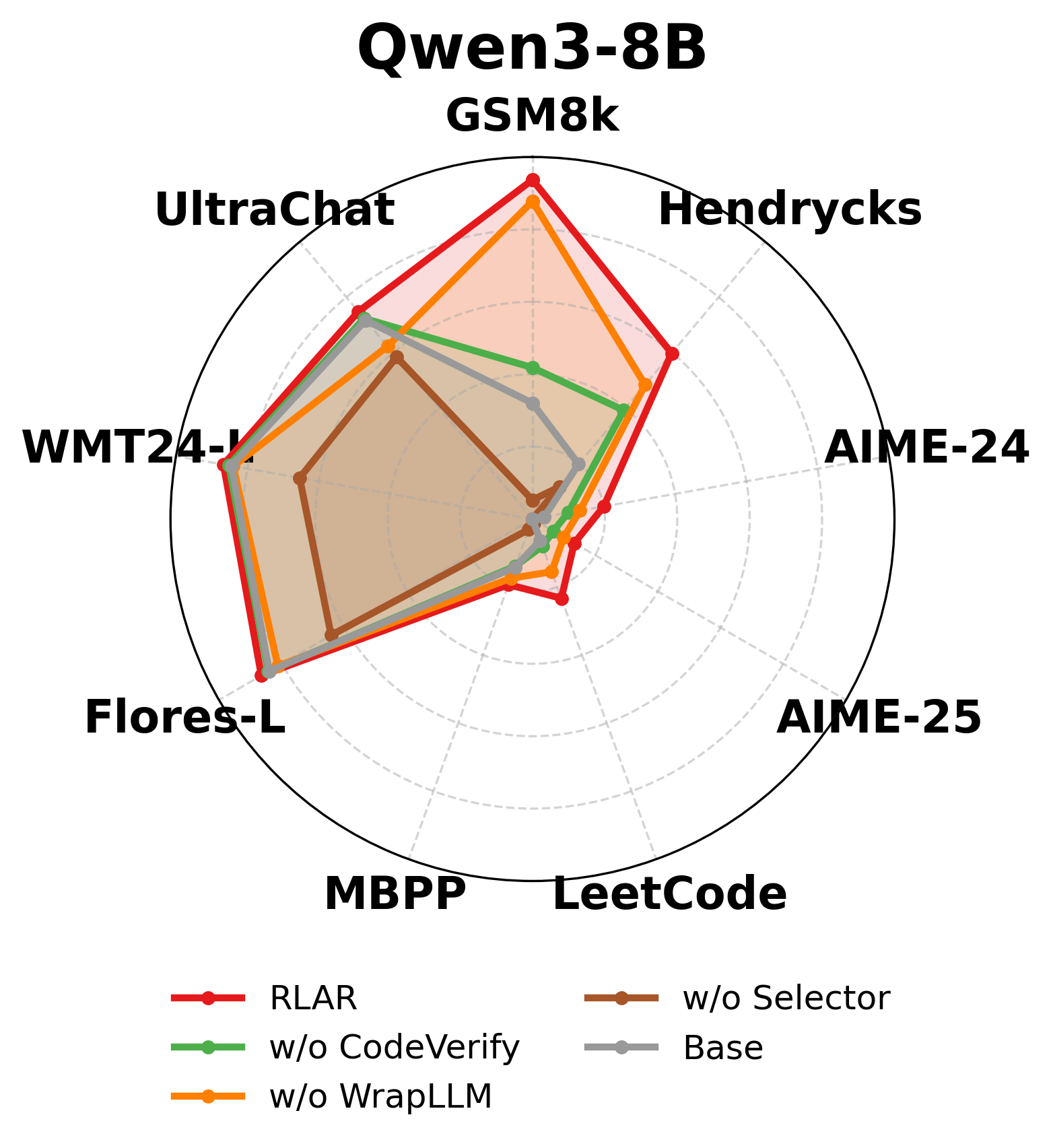}
         \caption{Qwen3-8B}
         \label{fig:abl_qwen}
     \end{subfigure}
     \caption{Ablation studies on Llama-3.1 and Qwen3.}
     \label{fig:ablation}
\end{figure}

To investigate the contributions of the actions within the \method{} framework, we conduct an ablation study with the following configurations: (1) \textbf{w/o CodeVerify} and (2)\textbf{ w/o WrapLLM}, which respectively disable the corresponding actions in the tool synthesis module; and (3) \textbf{w/o Selector}, where instead of immediate selection post-synthesis, a reward tool is randomly sampled only after all reward tools have been generated. The comparative results of these variants are visualized in Figure~\ref{fig:ablation}.

Removing the components of the tool synthesis module consistently results in degradation. Specifically, \textbf{without CodeVerify} designed tools, mathematical reasoning and coding task (\textsc{GSM8k, LeetCode}) drop significantly. \textbf{WrapLLM} is essential for chat and translation tasks. The above results confirm the necessity of query-specific reward designs, which is the core motivation of \method{}. 

\begin{table*}[t]
    \centering
    \small
        \begin{tabular}{lcccccc}
        \toprule
        \textbf{Method} & \textbf{Avg} & \textbf{Precision IF} & \textbf{Math} & \textbf{Safety} & \textbf{Factuality} & \textbf{Focus} \\ 
        \midrule
        SOTA        & 87.19 & 57.14 & 60.00 & 97.12 & 85.29 & 99.13 \\  \midrule
        \multicolumn{7}{l}{\textit {Agentic Selection within \method{} framework }} \\ 
        Random           & 78.39 & 40.48 & 60.00 & 89.42 & 75.49 & 90.43 \\
        GPT-4o    & 88.19 & 57.14 & 68.57 & 96.15 & 87.25 & 99.13 \\
        GLM-4.6   & 88.69 & 57.14 & 68.57 & 96.15 & 88.24 & 100.0 \\
        GPT-5.1          & \textbf{90.44} & \textbf{66.67} & \textbf{71.42} & 96.15 & \textbf{90.19} & \textbf{100.0} \\
        \midrule
        \multicolumn{7}{l}{\textit{ Logits Merging Methods }} \\ 
        mean@2           & 87.44 & 61.90 & 68.57 & 95.19 & 83.33 & 99.13 \\
        mean@10          & 77.89 & 47.62 & 57.14 & 84.62 & 76.47 & 90.43 \\
        mean@50          & 51.51 & 40.48 & 48.57 & 66.35 & 48.04 & 46.09 \\
        \midrule
        Theoretical Best & 93.47 & 76.19 & 77.14 & 97.12 & 95.10 & 100.0 \\
        \bottomrule
        \end{tabular}
    \caption{Comparison of Agentic Selection vs. Reward Merging on \textsc{RewardBench-v2}.}
    \label{tab:rewardbench_results}
    \vspace{-4mm}
\end{table*}

Moreover, the Random Reward Selection policy variant (\textbf{w/o Selector}) emphasizes that reward matching is important. The policy degrades on almost all testbenches. This confirms that the appropriate reward system design is crucial in the holistic procedure of RL training, and that our framework effectively synthesizes high-quality reward tools superior to random alternatives.

\subsection{Evaluating Reward Routing and Selection}

The core mechanism of the \method{} framework involves predicting and selecting the optimal reward function for a specific query triplet. 
As shown in the previous section, such a selection strategy is crucial for the performance. To demonstrate that \method{} achieves superior performance through the designed adaptive reward routing and selection, we conducted an analytical experiment using \textsc{RewardBench-v2}.

\noindent\textbf{Setup.} We utilized a randomly and uniformly sampled subset of 400 instances from the \textsc{RewardBench-v2} test set. Each instance comprises one preferred (chosen) response and three non-preferred (rejected) responses for a given prompt.\footnote{The ``Tie'' category is excluded due to the test's input-output format uniformity.}. For each instance, a reward model is required to score all four responses. An instance is considered a \textbf{pass} if the softmax-normalized reward score assigned to the preferred response exceeds a threshold of 0.5.

We leveraged the evaluation records of the top 50 classification models from the official leaderboard. RLAR is tasked with predicting the reward model most likely to pass a given instance, utilizing the query, candidate responses, and the models' metadata. For evaluation simplicity, reward models are presented to the agent as encapsulated tools. We compared the following approaches: \textbf{Backbone LLM Scaling}: Evaluating the impact of different agentic capabilities using GPT-4o, GLM-4.6, and GPT-5.1. \textbf{Reward Logits Merging}: Assessing the effect of ensemble methods by calculating the average scores of the top-$k$ models' predictions on each candidate responses (mean@$k$).

\noindent\textbf{Results and Discussion.} As shown in Table~\ref{tab:rewardbench_results}, simply merging reward model output logits (mean@$k$) does not effectively expand the performance ceiling; in fact, performance degrades significantly as more models are included (mean@50). In contrast, there is a clear positive correlation between the \textbf{agentic capability} of the backbone LLM and performance gains. These findings demonstrate that leveraging LLM reasoning to dynamically select reward models is a viable path for extending reward system design, approaching the theoretical performance limit of the available model pool.

We have further conducted the analysis on the WrapLLM search engine design. The average retrieved model repository position is 5.64 in the first page, showing the relatedness. \textbf{Further error analyses are included in Appendix}~\ref{sec:error_robustness}.

\section{Related Work}

\subsection{LLM RL Reward Designs}

In industry, training \textbf{discriminative} reward models~\citep{ouyangrlhf, deepseekr1, skyworkreward} is widely regarded as the most reliable approach for constructing a human preference oracle within reinforcement learning (RL) frameworks for LLM optimization. In addition, \textbf{generative} rewards extend the aforementioned task from classification to generation, and have demonstrated feasibility in mathematical domains~\citep{generativerm}, RLHF-based settings~\citep{ke-etal-2024-critiquellm, wang2024pandalm, zhu2025judgelm, autoj}, and can be integrated with advances in LLM reasoning, such as CritiqueGRPO~\citep{critiqueGRPO}.
With the rapid development of math reasoning and code generation, the design of \textbf{verifiable} rewards has attracted increasing attention. Binary rewards that can be verified through explicit rules have been shown to be more efficient in these domains~\citep{grpo, tulu3}. ~\citet{bleuberi} involves employing standard NLP metrics to guide instruction-following alignment as an extension of aforementioned approach.

\subsection{RL from AI Feedback}

RLAIF~\citep{lee2024rlaif} explores the development of reward models without extensive manual labeling of training data. Self-rewarding~\citep{selfreward} require the policy model to evaluate and discriminate its own generations. The LLM-as-a-judge~\citep{mtbench} paradigm employs a strong LLM to evaluate another LLM by means of a preceding evaluation prompt. RewardAgent~\citep{rewardagent} utilizes an LLM to combine pre-specified reward designs. These approaches inevitably embed strong human priors into reward design, either through the evaluation prompt or through the foundational reward specifications. In contrast to RewardAgent, our work extends both the design flexibility and the evaluation of reward design framework within an RL framework where reward is explicitly formulated (specifically GRPO rather than DPO).


\subsection{Dynamic Reward Assigning}

Recent research in integrating LLM with RL, particularly for reward shaping, has primarily focused on analyzing the agent's policy trace from prior steps to iteratively refine the reward function. ~\citep{afonso2025selfcorrectingrewardshapinglanguage} and~\citep{eager-2022} leverage the LLM's reasoning to guide reward weight pruning or analyze the trace to determine the appropriate reward shape design. Other methods, such as~\citep{logicRL} and~\citep{dynamic-rewarding-2024-emnlp} explore techniques like curriculum scheduling and adjusting the reward schedule via prompt hints. 
~\method{} diverges significantly by harnessing the LLM's capability to search the web and generate code, allowing it to directly \textbf{design entirely new rewards} rather than being limited to weight adjustments. ~\method{} is also flexible for \textbf{cross-domain optimization} problems, where reward designs differ substantially across various sub-domains, a challenge that existing single-task-focused methods do not fully address.

\section{Conclusion}

In this paper, we present \method{}, a novel agentic rewarding framework that designs query-specific reward functions for RL algorithm.
By integrating a self-evolving tool library with dynamic synthesis and selection, \method{} achieves superior performance across diverse domains, including math, code, translation and general chat.
Beyond empirical gains, \method{} demonstrates strong robustness against format and verbosity reward hacking. 
By significantly reducing the costs compared to RLAIF practices, \method{} establishes the ``LLM as Reward Designer'' paradigm.

\newpage

\section{Limitations}

We primarily validated \method{} on heterogeneous tasks in text forms. Due to the budget constraints, we did not extend the scope into multi-modal, audio tasks such as text-to-image generations. We believe this is a good exploration field for future works. There is still room for further analysis on the scalability of the \method{} framework.

In practice, some repository README would become out-dated when reporting (such as claiming to be state-of-the-art of that time). Though not directly caused by the design, \method{} is potentially vulnerable to readme hacking, as our assumption is that most of these repo readmes are trustworthy. We leave the development for developing more robust retrieval modules for future works.

Lastly, we focus on language models that are modeled as text classifiers. This is quite similar to practices in the industry, mainly aiming to save the computational cost of reward calculation. For generative reward models, our framework can support development on this basis; however, given the constraints of our experimental setup, we consider this to be outside the scope of the present work.


\bibliography{custom}

\appendix

\label{sec:appendix}

\section{Reproducibility Statement}

We have included the heterogeneous data construction process in Section~\ref{sec:task_design} and more details in Appendix. We described the RL training settings and experiment platforms in Appendix~\ref{apd:training_details} and Section~\ref{sec:experimental_setup}. The prompts involving the usage of LLM (primarily GPT-4.1) are filed in Appendix~\ref{apd:prompts}. The above materials are able to reproduce our work.

\section{Data Process Details} \label{apd:data}

\subsection{Detailed Introduction of Datasets}
\noindent\textbf{Translation (En-Fr, Fr-En)}: This task requires the LLM to translate between English and French (in our case, English to French and French to English). We use the dataset \texttt{aircrypto/English-French-Translations-Train-Large}~\citep{hf_enfr} from HuggingFace, which provides high-quality, paired sentence-level samples.

\noindent\textbf{Instruction Following}: Given specific requirements in the provided instructions, the LLM should respond accordingly. We use \texttt{tulu3-sft-reused-on-policy-8b}, part of the Tulu-3~\citep{tulu3} preference dataset, which contains generation pairs between different LLMs during the training of Llama‑3.1‑Tulu‑3‑8B.

\noindent\textbf{Multi-turn}: LLM respond to instructions with previous interaction histories. We pick \texttt{allenai-WildChat-1M-multiturn}~\citep{allenwilde}, a collection of 1M ChatGPT interaction logs from the wild. We select the English subset aimed at RLHF queries. 

\noindent\textbf{Summarization}: This task requires LLM to summarize over long documents into short abstracts. We pick \texttt{ccdv/govreport-summarization, ccdv/pubmed-summarization, ccdv/arxiv-summarization}\citep{cohan-etal-2018-discourse-summarizationdataset},  which includes different types of documents from arxiv articles to government reports. 

\noindent\textbf{Math}: We pick OpenAI \textsc{gsm8k}~\citep{gsm8k}, a classic dataset of grade‑school math problems designed to evaluate multi‑step reasoning. We choose not to use more complex math-reasoning datasets because our focus in this work is primarily on LLM text-generation tasks. Advanced math reasoning often requires specialized methodologies, such as tree‑search reasoning, which makes it unsuitable for single-pass direct generation.

\noindent\textbf{Conditional Generation}: The LLM should generate coherent text according to given constraints. In our setting, we task the LLM with filling in missing paragraphs in an essay or producing a complete essay based on an abstract outline. We use \texttt{qwedsacf/ivypanda-essays}~\citep{ivypanda}, a HuggingFace dataset repository containing long-form essays covering multiple disciplines sourced from the \textit{IvyPanda platform}\footnote{\texttt{https://ivypanda.com/}}.

\subsection{Data Filtering Prompt} ~\label{apd:prompt_for_data_filter}

\begin{tcolorbox}[promptbox, title={DATA\_FILTER\_PROMPT}]
\textbf{Input}: \texttt{task\_samples}, \texttt{quality\_standards}
\tcblower
You are given a set of task samples, each consisting of:
\begin{enumerate}
    \item \textbf{User Query} – the task or request made to the model.
    \item \textbf{Model Response} – the output given by the model.
\end{enumerate}

The samples may come from various task types, including: Translation, Summarization, Math problem solving, RLHF alignment, Conditional text generation, and Multi-turn dialogue.

\textbf{Your Goal}: Identify and select only the samples that did not meet quality standards based on:

\medskip
\textbf{A. Query Quality Issues}:
\begin{itemize}
    \item Ill-formed or incomplete queries; Ambiguous or misleading instructions.
    \item Irrelevant or off-topic requests; Grammatically broken or nonsensical input.
\end{itemize}

\textbf{B. Response Quality Issues}:
\begin{itemize}
    \item Incorrect or factually wrong answers; Incomplete responses.
    \item Poor language quality or incoherent writing.
    \item Hallucinations or made-up facts; Misinterpretation of the query.
\end{itemize}

\textbf{Instructions}:
\begin{enumerate}
    \item For each sample, examine both the query and response.
    \item Mark the sample as \textbf{"Fail"} if either the query quality or the response quality is below standard.
    \item Briefly explain why the sample fails, citing issues in query, response, or both.
    \item Output only the failing samples, in the format:
    \begin{tcolorbox}[colback=white, colframe=gray!20, size=minimal, left=2mm, top=1mm, bottom=1mm]
        \texttt{[Sample ID]} \\
        \texttt{Query: ...} \\
        \texttt{Response: ...} \\
        \texttt{Fail Reason: ...}
    \end{tcolorbox}
\end{enumerate}

\textit{Be strict in applying the criteria — even if only one side (query or response) is substandard, the sample should be considered as failing.}
\end{tcolorbox}
\section{Experiment Training Details}

\subsection{Training Configuration} \label{sec:training_config}
For the SFT setting, we tuned the base model on the training dataset for 2 epochs. All RL experiments were conducted using the GRPO~\citep{grpo} algorithm framework for 80 steps in total. Since Qwen3-8B is a reasoning model, we enforced a formatting constraint: the reward was set to zero if the reasoning chain failed to be properly encapsulated within `<think>' and `</think>' tags. We did not apply a length penalty during RL training, as generation length does not consistently correlate with task performance. More training details are filed in Appendix~\ref{apd:training_details}.

All experiments were performed on a cluster 8\(\times\)NVIDIA H100 GPUs (80GB), using a global batch size of 128 and mixed-precision (FP16) training. An auxiliary cluster of 8\(\times\)NVIDIA A100 GPUs (80GB) was dedicated to the deployment and inference of the reward models.





\subsection{Prompt for the LLM Judge}  
\label{apd:judge_prompt}
\begin{tcolorbox}[
    title = {LLM-as-a-judge Prompt}, 
    breakable, 
    colframe=blue!50!black, 
    colback=blue!10,
    fonttitle=\bfseries
]
\textbf{Input}: \texttt{prompt}, \texttt{candidate}, \texttt{reference}
\tcblower

You are an expert evaluator of language model outputs. You will receive:

\begin{enumerate}
    \item \textbf{Prompt:} The original instruction/task given to the model.
    \item \textbf{Candidate Response:} The model's output to be evaluated.
    \item \textbf{Reference Response:} A high-quality gold-standard or reference output.
\end{enumerate}

\textbf{Your task}:
\begin{itemize}
    \item Evaluate the quality of the \textit{Candidate Response} compared to the \textit{Reference Response} and in relation to the given \textit{Prompt}.
    \item Consider the task category (\textbf{translation}, \textbf{summarization}, \textbf{generation}, \textbf{infilling/cloze}, \textbf{conditional generation}, \textbf{math}, or \textbf{instruction following}) and adjust criteria accordingly.
    \item Score on a scale from \textbf{0 to 10} according to the rubric below.
    \item Output the score in the format \texttt{[[X]]} (where X is the integer) \textbf{once}, followed by a clear explanation.
\end{itemize}

\tcbline 

\subsubsection*{Evaluation Dimensions by Task Category}
\textit{(Use whichever are relevant to the given prompt)}

\begin{itemize}
    \item \textbf{Translation}: Accuracy, fidelity, fluency, grammar, style.
    \item \textbf{Summarization}: Coverage, factual faithfulness, conciseness, coherence.
    \item \textbf{Generation}: Relevance, originality, creativity, coherence, style.
    \item \textbf{Infilling/Cloze}: Correctness of missing content, contextual fit, fluency.
    \item \textbf{Math/Reasoning}: Correctness of logic, clarity, rigor of explanation.
    \item \textbf{Instruction Following}: Alignment with intent, completeness.
\end{itemize}

\tcbline

\subsubsection*{Scoring Rubric (0–10)}
\begin{small}
\begin{description}
    \item[10:] \textbf{Perfect}. Fully correct, faithful, and clear. No errors.
    \item[8--9:] \textbf{Very Good}. Minor style issues or tiny, overlookable omissions.
    \item[6--7:] \textbf{Good/Fair}. Mostly correct but with notable small issues or some loss of clarity.
    \item[4--5:] \textbf{Poor/Borderline}. Mix of correct/incorrect; noticeable gaps; low reliability.
    \item[1--3:] \textbf{Very Poor}. Incoherent, irrelevant, or almost entirely wrong.
    \item[0:] \textbf{No meaningful output} or completely unrelated.
\end{description}
\end{small}

\tcbline

\subsubsection*{Output Format}
Respond with:
\begin{tcolorbox}[colback=white, colframe=gray!30, size=minimal, left=2mm, top=2mm, bottom=2mm]
\texttt{[[X]]} \\
\texttt{Explanation:[Detailed explanation within 50 words, citing criteria.]}
\end{tcolorbox}

\tcbline

\textbf{[Prompt]} \\
\texttt{\{prompt\}}

\medskip
\textbf{[Candidate Response]} \\
\texttt{\{candidate\}}

\medskip
\textbf{[Reference Response]} \\
\texttt{\{reference\}}

\end{tcolorbox}

\subsection{Hyperparameters Details for RL Training} \label{apd:training_details}

We use the volcano engine reinforcement learning for LLMs framework, \textsc{verl}~\citep{verlengine}. We validate the implementation of the framework run all our RL experiments based on it. Below is the hyperparameters for all our experiments and we use the same set of hyperparameters for all experiments.

\begin{table}[htbp]
\centering
\small
\begin{tabularx}{\columnwidth}{Xr}
\toprule
\textbf{Category / Parameter} & \textbf{Value} \\
\midrule
\rowcolor[gray]{0.9} \multicolumn{2}{l}{\textit{Training \& Data}} \\
Total Epochs & 5 \\
Global Train Batch Size & 256 \\
Max Prompt Length & 10,000 \\
Max Response Length & 5,000 \\
Truncation Strategy & 'error' \\
Filter Overlong Prompts & True \\
\midrule
\rowcolor[gray]{0.9} \multicolumn{2}{l}{\textit{Algorithm (GRPO/PPO)}} \\
Advantage Estimator & GRPO \\
Learning Rate & $1 \times 10^{-6}$ \\
PPO Mini-batch Size & 16 \\
KL Coefficient & 0.0001 \\
KL Loss Type & low\_var\_kl \\
Entropy Coefficient & 0 \\
KL in Reward & False \\
\midrule
\rowcolor[gray]{0.9} \multicolumn{2}{l}{\textit{Rollout (vLLM Engine)}} \\
Rollout Samples ($n$) & 8 \\
Max Batched Tokens & 65,536 \\
GPU Memory Utilization & 0.7 \\
Prompt/Response Length & 10k / 5k \\
\midrule
\rowcolor[gray]{0.9} \multicolumn{2}{l}{\textit{Parallelism \& Infrastructure}} \\
Nodes / GPUs per Node & 1 / 8 \\
Tensor Parallel (TP) Size & 2 \\
Pipeline Parallel (PP) Size & 4 \\
Gradient Checkpointing & True \\
Micro Batch Size (Actor) & 1 \\
\bottomrule
\end{tabularx}
\caption{Hyperparameter settings for GRPO training with Qwen3-8B.}
\label{tab:hyperparams}

\end{table}

\begin{table}[t]
\centering
\small
\begin{tabularx}{\columnwidth}{Xr}
\toprule
\textbf{Category / Parameter} & \textbf{Value} \\
\midrule
\rowcolor[gray]{0.9} \multicolumn{2}{l}{\textit{Training \& Data}} \\
Total Epochs & 5 \\
Global Train Batch Size & 256 \\
Max Prompt Length & 10,000 \\
Max Response Length & 5,000 \\
Truncation Strategy & 'error' \\
Filter Overlong Prompts & True \\
\midrule
\rowcolor[gray]{0.9} \multicolumn{2}{l}{\textit{Algorithm (GRPO)}} \\
Advantage Estimator & GRPO \\
Learning Rate & $1 \times 10^{-6}$ \\
PPO Mini-batch Size & 16 \\
KL Coefficient & 0.0001 \\
KL Loss Type & low\_var\_kl \\
Entropy Coefficient & 0 \\
KL in Reward & False \\
\midrule
\rowcolor[gray]{0.9} \multicolumn{2}{l}{\textit{Rollout (vLLM Engine)}} \\
Rollout Samples ($n$) & 8 \\
Max Batched Tokens & 65,536 \\
GPU Memory Utilization & 0.7 \\
Prompt/Response Length & 10k / 5k \\
\midrule
\rowcolor[gray]{0.9} \multicolumn{2}{l}{\textit{Parallelism \& Infrastructure}} \\
Nodes / GPUs per Node & 1 / 8 \\
Tensor Parallel (TP) Size & 2 \\
Pipeline Parallel (PP) Size & 4 \\
Gradient Checkpointing & True \\
Micro Batch Size (Actor) & 1 \\
\bottomrule
\end{tabularx}
\caption{Hyperparameter settings for GRPO training with Llama-3.1-8B.}
\label{tab:llama_hyperparams}

\end{table}

The other hyper-parameters, such as optimizer $\beta$, are set default to the framework trainer configurations from \url{https://github.com/volcengine/verl/tree/main/verl/trainer/config}.

\section{Prompt Details} \label{apd:prompts}

\subsection{Prompt for Task Decomposition}  \label{apd:prompt_task_decpm}

\begin{tcolorbox}[title = {Task\_Decomposition\_Prompt}, breakable, colframe=blue!50!black, colback=blue!10]
\textbf{Input}: original\_task
\tcblower

Please break down the following generative task into a combination of several basic generative tasks:  

Basic task list:
1. Controlled generation: Generate coherent natural language text that meets certain given conditions. Best for simple, clear tasks; complex writing should be split into smaller steps like planning and cloze generation.

2. Translation: Generate a corresponding text in another natural language from a text in one natural language.  

3. Text summarization: Summarize the given text, retaining the main information.  

4. Question answering: Provide appropriate answers based on background information and question requests provided by the user.  

5. Paraphrasing: Modify the provided text into a different form of expression that meets the given rewriting requirements.  

6. Cloze generation: Given a continuous piece of text with missing parts, generate appropriate text for the missing positions so that the original text becomes complete, coherent, and consistent.  

7. Planning generation: Plan a high-level outline in order to accomplish a relatively complex generative task, such as creating a chapter list, designing character traits, designing scripts, or designing a timeline.  

8. Code: Generate executable code that meets the specified requirements, or supplement or revise code according to the given requirements. The defining criterion for this task is that the output is primarily code.

Decomposition goal:

- Break down the complex generative task provided by the user into a list composed of the above basic tasks according to its logical steps.  
- Steps should be arranged in execution order, and the description should start from the original input form and proceed until the task is completed.  

- Each step must clearly specify the “basic task type” and the execution content of that step.  

- If the task does not need to be broken down, provide a single-step basic task and rewrite its description into a clearer instruction that aligns with the type of task in the basic task list.

Output format requirements:

- List the decomposition results step-by-step (step number + basic task type name + specific execution description). 

- Enclose the final result within $<$Result$>$ ... $<\backslash$Result $>$  tags.  

Below is an example:

[Example Start]

Task to be decomposed: Please provide an English summary for the following Chinese document.

Decomposition result:  

1. Translation: Please translate the following Chinese document into an English document.  

2. Text summarization: Please summarize the given English document, and ensure the summary does not exceed 200 words.  

[Example End]  

Now, perform the above decomposition process on the given question (or task description) below, and write the final decomposition result within $<$Result$>$ ... $<\backslash$Result $>$  tags.

\{original\_task\}

\end{tcolorbox}

\subsection{Prompt Details for Reward Model Choice}

\subsubsection{Search API Tool} \label{apd:tool_snippet}

\begin{lstlisting}[style=mypython]
{
    "type": "function",
    "function": {
        "name": "search_serper_engine",
        "description": "Performs a Google search using the Serper API restricted to finding Hugging Face model checkpoints. Use this tool only to look up Hugging Face checkpoint URLs, model pages, or related information. Short queries work best. Reward model might be confusing with base models or chat models",
        "parameters": {
            "type": "object",
            "properties": {
                "query": {
                    "type": "string",
                    "description": "The search query for Hugging Face checkpoints, e.g., model names or keywords to locate on huggingface.co."
                }
            },
            "required": ["query"]
        }
    }
}
\end{lstlisting}

\subsubsection{Prompt for search results filtration}  
\label{apd:filter_prompt}
\begin{tcolorbox}[title = {Search Results Filtration}, breakable, colframe=blue!50!black, colback=blue!10]
\textbf{Input}: \texttt{original\_task}
\tcblower

You are given a list of search engine results with position IDs. 

Your task is to filter them according to the following rules:

\begin{enumerate}
    \item \textbf{Identify Reward Models:} 
    \begin{itemize}
        \item Keep only results that are \textbf{reward model} links.
        \item Reward models often have model names containing keywords like \texttt{-Reward-} or \texttt{-RM-}.
        \item Discard results for base models (\texttt{-Base}), instruct models (\texttt{-Instruct}), or chat models (\texttt{-Chat}).
        \item If a model name has none of these hints, and it's unclear whether it is a reward model, discard it.
    \end{itemize}

    \item \textbf{Hugging Face Model Repositories Only:}
    \begin{itemize}
        \item Keep only links pointing to \textbf{Hugging Face model repositories}.
        \item Discard datasets, research papers, blog posts, or other non-model content.
    \end{itemize}

    \item \textbf{Score Output Format only:}
    \begin{itemize}
        \item Regression models only, in other words, models that output a score (e.g., 0--1) rather than generating text.
    \end{itemize}
\end{enumerate}

Directly discard those items that violate rule 1, 2, or 3 and keep the rest. Output the remaining items in a list using their original position IDs like \texttt{[0, 1, 3, 5, ...]}. If none of the items are left, output an empty list \texttt{[]}.

\bigskip
\{results\}

\end{tcolorbox}

\subsubsection{Prompt for search results Reranking}  
\label{apd:rerank_prompt}

\begin{tcolorbox}[title = {Search Results Reranking}, breakable, colframe=blue!50!black, colback=blue!10]
\textbf{Input}: \texttt{original\_task}
\tcblower

You are given a list of search engine results with position IDs.  
Your task is to filter and rank them according to the following rules:

\begin{enumerate}
    \item \textbf{Identify Reward Models:}  
    \begin{itemize}
        \item Keep only results that are \textbf{reward model} links.  
        \item Reward models often have model names containing keywords like \texttt{-Reward-} or \texttt{-RM-}.  
        \item Discard results for base models (\texttt{-Base}), instruct models (\texttt{-Instruct}), or chat models (\texttt{-Chat}).  
        \item If a model name has none of these hints, and it's unclear whether it is a reward model, discard it.
    \end{itemize}

    \item \textbf{Hugging Face Model Repositories Only:}  
    \begin{itemize}
        \item Keep only links pointing to \textbf{Hugging Face model repositories}.  
        \item Discard datasets, research papers, blog posts, or other non-model content.
    \end{itemize}

    \item \textbf{Score Output Format only:}
    \begin{itemize}
        \item Regression models only; in other words, models that output a score (e.g., 0--1) rather than generating text.
    \end{itemize}
\end{enumerate}

Directly discard those items that violate rule 1, 2, or 3 and keep the rest. Output the remaining items in a list using their original position IDs, sorted by relevance, like \texttt{[0, 1, 3, 5, ...]}. If none of the items are left, output an empty list \texttt{[]}.

\bigskip
\{results\}

\end{tcolorbox}

\subsubsection{Prompt for search results LLM-based Reward Model Implementation}  
\label{apd:implement_prompt}

\begin{tcolorbox}[title = {Reward Tool Implementation}, breakable, colframe=blue!50!black, colback=blue!10]
\textbf{Input}: original\_task
\tcblower

Implement a python script for launching a reward model according to the following informative scripts. The model local checkpoint is \{model\_local\_dir\}. The cuda device for the model is "\{cuda\_device\}". You should write a function, that support input parameter:
- prompt: str, instruction or context conditions
- response: str, the text need to be evaluated
- reference: str, some reference answer/response for the above prompt

Your implementation are free to use the packages mentioned in the scripts. Name the calculation function starting with "compute\_", such as "def compute\_XXX(...)" where XXX should be the reward model name or related abbreviation. Make sure the model checkpoint is loaded precisely once in the script. Format your output enclosed within "python $\backslash$ n xxxx $\backslash$n". Also, additionally print the calculation funciton after four sharp marks \#\#\#\#, such as "\#\#\#\# def compute\_xxx(...)" in the end of your output (outside the python script).

\{scripts\}

[your implementation]

\end{tcolorbox}

\subsection{Prompt for Code-Agent workflow}

\subsubsection{Plan} \label{sec:code-agent-plan}

\begin{tcolorbox}[promptbox, title={LIST\_TASK\_PROMPT}]
\textbf{Input}: \texttt{\{task\}}
\tcblower
You are an expert in designing reward models and evaluation metrics for the \textbf{\{task\}} task.  
Your goal is to list \textbf{3–5 possible reward model or evaluation metric choices} for this task, drawing from the following two categories:  

\begin{enumerate}
    \item \textbf{Rule-based} – Explicit rules (e.g., exact match with reference output, length constraints) used directly as rewards.  
    \item \textbf{Metric-based} – Standard NLP metrics (e.g., BLEU, ROUGE, METEOR) used to evaluate and reward generated results.  
\end{enumerate}

\textbf{Output formatting requirements}:  
\begin{itemize}
    \item Place your results \textbf{after four hash marks (\texttt{\#\#\#\#})}.
    \item For \textbf{each choice}, indicate its \textbf{category} and \textbf{name}, using the format:
    \begin{quote}
        \texttt{\#\#\#\# <Category>/<Name>: <Brief description>}
    \end{quote}
    \item Use a \textbf{new line} for each choice.
\end{itemize}

\textbf{Example}:  
\begin{tcolorbox}[colback=white, colframe=gray!30, size=small]
\texttt{\#\#\#\# Metric-based/BLEU: Measures the n-gram overlap between generated output and reference text.}\\
\texttt{\#\#\#\# Rule-based/Length: Rewards outputs within the target length range for conciseness.}
\end{tcolorbox}
\end{tcolorbox}

\subsubsection{Write} \label{sec:code-agent-write}

\begin{lstlisting}[style=mypython]
WRITE_CODE_PROPMT = """Implement the following metric according to description using python. You are free to use packages. You should write a function begin with 'compute_xxx' where xxx is the name of the metric. The function accepts:
- prompt: the instruction to the prompt
- candidate_response: the candidate response to be evaluated by the metric
- reference_response: the reference answer for the prompt
You should directly return a scaler score.

Output the python code in ```python\n xxxx\n```. And list the requirements within `````` use requirements.txt style. 
  
{metric description}
"""
\end{lstlisting}

\subsection{Prompts for WrapLLM and CodeVerify} \label{apd:prompt_wrap_code}

\begin{tcolorbox}[title={\texttt{TASK\_CLS\_PROMPT}}, breakable, colframe=blue!50!black, colback=blue!5]
You will be given a \texttt{question} and an \texttt{answer} from a language model interaction.  
Your job is to determine the main type of task being performed. Examples include but are not limited to: translation, summarization, question answering, code generation, math solving, creative writing, RLHF alignment, explanation, classification, etc.  
You do not need to list all possible task types — instead, use your judgment to give the most fitting label for this specific instance.  
\textbf{Rules:}  
\begin{enumerate}
    \item Output only the task type as a concise label (maximum \textbf{three words}).
    \item Do not include any extra text, punctuation, or explanation.
    \item If uncertain, choose the closest-fitting description.
\end{enumerate}

Provide your output results after four sharp marks \texttt{\#\#\#\#}, such as \texttt{"\#\#\#\# Translation"}.

\textbf{Input example:}  
\begin{quote}
\texttt{"question": "You are a helpful assistant that translates English sentences to French... [English Input]\textbackslash n Michael Gill formed The Murder Mile... [French Output]\textbackslash n",} \\
\texttt{"answer": "Michael Gill a formé The Murder Mile..." }
\end{quote}
\textbf{Output:}  
\texttt{\#\#\#\# English--French Translation}

Now classify the given instance according to these rules.
\end{tcolorbox}

\begin{tcolorbox}[title={\texttt{TASK\_DECOMP\_PROMPT}}, breakable, colframe=blue!50!black, colback=blue!5]
Please break down the following generative task into a combination of several basic generative tasks:

\textbf{Basic task list:}
\begin{enumerate}
    \item \textbf{Controlled generation:} Generate coherent natural language text that meets certain conditions.
    \item \textbf{Translation:} Generate corresponding text in another natural language.
    \item \textbf{Text summarization:} Summarize the given text.
    \item \textbf{Question answering:} Provide appropriate answers based on background info.
    \item \textbf{Paraphrasing:} Modify text into a different form of expression.
    \item \textbf{Cloze generation:} Fill in missing parts of a text.
    \item \textbf{Planning generation:} Plan a high-level outline (chapter lists, scripts).
    \item \textbf{Code:} Generate or revise executable code.
\end{enumerate}

\textbf{Decomposition goal:}
\begin{itemize}
    \item Break down the complex task into a list of the above basic tasks.
    \item Steps should be in execution order.
    \item Each step must specify the ``basic task type'' and execution content.
\end{itemize}

\textbf{Output format requirements:}
\begin{itemize}
    \item List results step-by-step (step number + type + description).
    \item Enclose the final result within \texttt{<Result> ... </Result>} tags.
\end{itemize}

\bigskip
\texttt{\{original\_task\}}
\end{tcolorbox}

\begin{tcolorbox}[title={\texttt{LIST\_TASK\_PROMPT}}, breakable, colframe=blue!50!black, colback=blue!5]
You are an expert in designing reward models and evaluation metrics for the \textbf{\texttt{\{task\}}} task.  
Your goal is to list \textbf{3--5 possible reward model or evaluation metric choices}, drawing from:  
\begin{enumerate}
    \item \textbf{Rule-based} -- Explicit rules (e.g., exact match, length constraints).
    \item \textbf{Metric-based} -- Standard NLP metrics (e.g., BLEU, ROUGE).
\end{enumerate}

\textbf{Output formatting requirements:}  
\begin{itemize}
    \item Place your results \textbf{after four hash marks (\texttt{\#\#\#\#})}.
    \item Format: \texttt{\#\#\#\# <Category>/<Name>: <Brief description>}
    \item Use a \textbf{new line} for each choice.
\end{itemize}
\end{tcolorbox}

\begin{tcolorbox}[title={\texttt{WRITE\_CODE\_PROPMT}}, breakable, colframe=blue!50!black, colback=blue!5]
Implement the following metric according to description using python. Write a function begin with \texttt{'compute\_xxx'}. The function accepts: \texttt{prompt}, \texttt{candidate\_response}, \texttt{reference\_response}. Return a scalar score.

Output the python code in \texttt{```python ... ```}. List requirements in \texttt{requirements.txt} style.

\medskip
\texttt{[metric description]}
\end{tcolorbox}

\begin{tcolorbox}[title={\texttt{WRAP\_TOOL\_PROMPT}}, breakable, colframe=blue!50!black, colback=blue!5]
Write a short description of the following reward function. Briefly explain what it calculates and how to use it for NLP evaluation.

\textbf{NAME:} \texttt{\{name\}} \\
\textbf{CODE:} \\
\texttt{```python} \\
\texttt{\{code\}} \\
\texttt{```}

Write your description after \texttt{\#\#\#\#}, e.g., \texttt{"\#\#\#\# This function calculates..."}.
\end{tcolorbox}

\begin{tcolorbox}[title={\texttt{METRIC\_DESIGN\_PROMPT}}, breakable, colframe=blue!50!black, colback=blue!5]
You are an expert in designing evaluation metrics with python. Implement \texttt{\{metric\}}. 

\textbf{Info:} \texttt{\{info\}}

Write a function \texttt{def compute\_XXX(prompt, candidate\_response, reference\_response)} that returns the score. Enclose in \texttt{```python ... ```}.
\end{tcolorbox}

\begin{tcolorbox}[title={\texttt{REWARD\_MODEL\_DESIGN\_PROMPT}}, breakable, colframe=blue!50!black, colback=blue!5]
Expert in reward models. Write a script to calculate reward scores. 
Function: \texttt{def compute\_XXX(prompt, response, reference)}. 
Device: \texttt{"cuda:0"}. Load model precisely once.

\textbf{README.md:} \texttt{\{readme\}} \\
\textbf{Path:} \texttt{\{model\_path\}}
\end{tcolorbox}

\begin{tcolorbox}[title={\texttt{RERANK\_PROMPT}}, breakable, colframe=blue!50!black, colback=blue!5]
Filter search results based on:
\begin{enumerate}
    \item \textbf{Identify Reward Models:} Keep \texttt{-Reward-} or \texttt{-RM-}. Discard \texttt{-Base}, \texttt{-Instruct}, \texttt{-Chat}.
    \item \textbf{Hugging Face Only:} Model repositories only. Discard datasets/papers.
    \item \textbf{Score Output:} Regression models (e.g., 0--1).
\end{enumerate}
Output original IDs as a list: \texttt{[0, 1, 3]}.

\medskip
\texttt{\{results\}}
\end{tcolorbox}

\begin{tcolorbox}[title={\texttt{LLMRM\_IMPLEMENT\_CODE}}, breakable, colframe=blue!50!black, colback=blue!5]
Implement python script for reward model. \\
\textbf{Checkpoint:} \texttt{\{model\_local\_dir\}} \\
\textbf{Device:} \texttt{"\{cuda\_device\}"} \\
\textbf{Function:} \texttt{def compute\_XXX(prompt, response, reference)}.

Format: \texttt{```python ... ```}. Print function signature after \texttt{\#\#\#\#} at the end.

\medskip
\texttt{\{scripts\}}
\end{tcolorbox}

\begin{tcolorbox}[title={\texttt{SELECT\_INFORMATIVE\_FILES}}, breakable, colframe=blue!50!black, colback=blue!5]
Identify informative files from HF repo: \\
\texttt{\{file\_list\}}

Output list: \texttt{["file1.py", "README.md"]}. No extra text.
\end{tcolorbox}
\section{LLM usage in this paper}

Large Language Models (LLMs) were used in the preparation of this work as a general-purpose assistance tool. Specifically, LLMs were employed in the following ways:
\begin{itemize}
    \item \textbf{Translation Assistance}: Converting expressions and sentences from the author’s native language into English.
    \item \textbf{Language Polishing and Grammar Revision}: Improving clarity, fluency, and grammatical correctness of the text, and ensuring that phrasing is natural in academic English.
    \item \textbf{Draft Review and Critique}: Providing feedback on drafts, including identifying unclear passages, suggesting improvements in structure, and flagging potential ambiguities.
\end{itemize}

LLMs were not used for generating original research ideas, performing data analysis, or writing substantive technical content. All core research contributions, results, and argumentative structure were developed by the authors. The role of LLMs was limited to translation, linguistic polishing, and non-substantive editorial suggestions to improve presentation.

\section{Generated Tools}




\onecolumn

\begin{longtable}{|l|l|l|l|}
\caption{A list of the generated reward function tool names by our code-agent.} \label{tab:tool_name_list} \\
\hline 
\textbf{Type} & \textbf{Metric} & \textbf{Type} & \textbf{Metric} \\
\hline
\endfirsthead

\hline
\textbf{Type} & \textbf{Metric} & \textbf{Type} & \textbf{Metric} \\
\hline
\endhead

rule\_based & \texttt{Forbidden\_Words} & rule\_based & \texttt{Stepwise\_Completeness} \\
rule\_based & \texttt{Prompt\_Adherence} & rule\_based & \texttt{Length} \\
rule\_based & \texttt{Numeric\_Accuracy} & rule\_based & \texttt{Exact\_Template\_Match} \\
rule\_based & \texttt{Novelty\_Penalty} & rule\_based & \texttt{Contradiction\_Detection} \\
rule\_based & \texttt{Disallowed\_Phrase\_Penalty} & rule\_based & \texttt{Exact\_Output\_Match} \\
rule\_based & \texttt{No\_Unsupported\_Claims} & rule\_based & \texttt{Reference\_Match} \\
rule\_based & \texttt{Exact\_Answer\_Match} & rule\_based & \texttt{Named\_Entity\_Preservation} \\
rule\_based & \texttt{Unit\_Consistency} & rule\_based & \texttt{Keyword\_Presence} \\
rule\_based & \texttt{Minimal\_Edit\_Distance} & rule\_based & \texttt{Thesis\_Inclusion} \\
rule\_based & \texttt{Mandatory\_Content\_Inclusion} & rule\_based & \texttt{Scientific\_Claims\_Match} \\
rule\_based & \texttt{Pronounceability} & rule\_based & \texttt{Position\_Sensitivity} \\
rule\_based & \texttt{Section\_Coverage} & rule\_based & \texttt{Entity\_Presence} \\
rule\_based & \texttt{Answer\_Type\_Match} & rule\_based & \texttt{Stepwise\_Correctness} \\
rule\_based & \texttt{Terminology\_Accuracy} & rule\_based & \texttt{Diversity\_Score} \\
rule\_based & \texttt{Forbidden\_Content} & rule\_based & \texttt{Fact\_Match} \\
rule\_based & \texttt{Forbidden\_Phrase\_Detection} & rule\_based & \texttt{No\_Information\_Leakage} \\
rule\_based & \texttt{Annotation\_Completeness} & rule\_based & \texttt{Grammar\_and\_Spelling\_Accuracy} \\
rule\_based & \texttt{Clarity\_Constraint} & rule\_based & \texttt{Answer\_Presence} \\
rule\_based & \texttt{No\_Overlap\_with\_Input} & rule\_based & \texttt{No\_Syntax\_Errors} \\
rule\_based & \texttt{Numeric\_Tolerance} & rule\_based & \texttt{Edit\_Distance} \\
rule\_based & \texttt{Keyword\_Coverage} & rule\_based & \texttt{No\_Repetition} \\
rule\_based & \texttt{Length\_Ratio} & rule\_based & \texttt{One-Hot\_Accuracy} \\
rule\_based & \texttt{Novelty} & rule\_based & \texttt{Exact\_Match} \\
rule\_based & \texttt{Pattern\_Compliance} & rule\_based & \texttt{Step\_Match} \\
rule\_based & \texttt{Syntax\_Validity} & rule\_based & \texttt{Format\_Compliance} \\
rule\_based & \texttt{Allowed\_Vocabulary} & rule\_based & \texttt{Entity\_Overlap} \\
rule\_based & \texttt{Explicit\_Irrelevance} & rule\_based & \texttt{Accuracy} \\
rule\_based & \texttt{Coverage\_of\_Key\_Points} & rule\_based & \texttt{Section\_Presence} \\
rule\_based & \texttt{Clarity} & rule\_based & \texttt{Test\_Case\_Pass\_Rate} \\
rule\_based & \texttt{Dictionary\_Filtering} & rule\_based & \texttt{Length\_Expansion} \\
rule\_based & \texttt{Content\_Inclusion} & rule\_based & \texttt{Error\_Pattern\_Removal} \\
rule\_based & \texttt{Plagiarism\_Check} & rule\_based & \texttt{Functionality\_Test} \\
rule\_based & \texttt{Politeness\_Constraint} & rule\_based & \texttt{Formatting\_Compliance} \\
rule\_based & \texttt{Exact\_Test\_Case\_Pass} & rule\_based & \texttt{Key\_Information\_Coverage} \\
rule\_based & \texttt{Genre-Adherence} & rule\_based & \texttt{Passes\_Unit\_Tests} \\
rule\_based & \texttt{Exact\_Step\_Match} & rule\_based & \texttt{Exact\_Keyword\_Match} \\
rule\_based & \texttt{Required\_Field\_Inclusion} & rule\_based & \texttt{Attribute\_Coverage} \\
rule\_based & \texttt{Valid\_Vocabulary} & rule\_based & \texttt{Medical\_Term\_Coverage} \\
rule\_based & \texttt{Keyword\_Absence} & rule\_based & \texttt{Required\_Component\_Presence} \\
rule\_based & \texttt{Final\_Answer\_Correctness} & rule\_based & \texttt{Keyword\_Inclusion} \\
rule\_based & \texttt{Structure\_Compliance} & rule\_based & \texttt{Step\_Consistency} \\
rule\_based & \texttt{Readability} & rule\_based & \texttt{No-Answer\_Accuracy} \\
rule\_based & \texttt{Length\_Constraint} & rule\_based & \texttt{Error\_Reduction} \\
rule\_based & \texttt{Answer\_Type\_Mismatch} & rule\_based & \texttt{Thesis\_Presence} \\
rule\_based & \texttt{Case-Insensitive\_Match} & rule\_based & \texttt{Topic\_Divergence} \\
rule\_based & \texttt{Exact\_Numeric\_Match} & rule\_based & \texttt{Originality-Penalty} \\
rule\_based & \texttt{Keyword\_Exclusion} & rule\_based & \texttt{Structure} \\
rule\_based & \texttt{Format\_Consistency} & rule\_based & \texttt{Required\_Elements} \\
rule\_based & \texttt{Reference\_Citation} & rule\_based & \texttt{Instruction\_Match} \\
rule\_based & \texttt{Key\_Concepts\_Inclusion} & rule\_based & \texttt{Stepwise\_Solution\_Match} \\
rule\_based & \texttt{Fact\_Consistency} & rule\_based & \texttt{Step\_Count\_Constraint} \\
nlp\_metric & \texttt{F1\_Score} & nlp\_metric & \texttt{METEOR} \\
nlp\_metric & \texttt{ROUGE} & nlp\_metric & \texttt{GLEU} \\
nlp\_metric & \texttt{BERTScore} & nlp\_metric & \texttt{M\^{}2\_Score} \\
nlp\_metric & \texttt{chrF} & nlp\_metric & \texttt{ROUGE-L} \\
nlp\_metric & \texttt{Levenshtein\_Distance} & nlp\_metric & \texttt{BLEU} \\
nlp\_metric & \texttt{Distinct-n} & nlp\_metric & \texttt{CodeBLEU} \\
model\_based & \texttt{Content\_Novelty\_Score} & model\_based & \texttt{Negative\_Relevance\_Score} \\
model\_based & \texttt{Topic\_Classifier} & model\_based & \texttt{Perplexity} \\
\hline
\end{longtable}

\clearpage
\twocolumn

\begin{table}[t]
    \centering
    \resizebox{\linewidth}{!}{
    \begin{tabular}{l|l}
    \hline
\textbf{Source} & \textbf{Repo Name} \\ \midrule
\citep{skyworkreward}   & Skywork/Skywork-Reward-V2-Llama-3.1-8B \\
\citep{skyworkreward}         & Skywork/Skywork-Reward-V2-Qwen3-8B \\
\citep{skyworkreward}         & Skywork/Skywork-Reward-V2-Llama-3.2-3B \\
\citep{skyworkreward}         & Skywork/Skywork-Reward-V2-Qwen3-4B \\
\citep{seedX} & ByteDance-Seed/Seed-X-RM-7B\\
                            & OpenAssistant/reward-model-deberta-v3-base \\
\citep{yang2024gpt2helpful}         & Ray2333/gpt2-large-helpful-reward\_model \\
\citep{nicholas22rewardreward}         & nicholasKluge/RewardModel \\
\citep{cai2024internlm2}         & internlm/internlm2-1\_8b-reward \\
\citep{malik2025rewardbench2advancingreward}         & allenai/Llama-3.1-8B-Base-RM-RB2 \\ \bottomrule
    \end{tabular}
    }
    \caption{Successfully Top@9 deployed LLM-based reward models.}
    \label{tab:llm-rm}
\end{table}


\section{Additional Analysis}


\subsection{Error and Robustness Analysis} \label{sec:error_robustness}


\begin{table}[t]
    \centering
    \small
    \begin{tabular}{lc}
      \toprule
      \textbf{Task Type} & \textbf{Unmatched(\%)}\\
      \midrule
      Infilling & $47.4$ \\
      Essay Generation & $43.8$ \\
      Multi-Turn & $8.8$ \\
      \bottomrule
    \end{tabular}
    \caption{Web Retrieval Page Rank}
    \label{tab:search_robustness}
\end{table}

We conducted an error analysis by counting the task types of instructions for which the Web-Agent could not find a specialized reward model (unmatched conditions). The breakdown in Table \ref{tab:unmatched_errors} shows that the majority of unfound instructions originate from essay infilling/generation tasks. Specifically, there is currently no corresponding reward model explicitly trained for the these two task domains, which accounts for the high unmatched ratio in these categories. Notably, When a specialized tool is unmatched, ~\method{} defaults to using a generic, default LLM-based reward model (\texttt{skywork-llama}).

\begin{table}[t]
    \centering
    \small
    \begin{tabular}{lc}
      \toprule
      \textbf{Category} & \textbf{Avg Pos}\\
      \midrule
      Summ & $7.17$ \\
      Translation & $2.36$ \\
      RLHF & $5.03$\\
      Multi-Turn & $7.61$ \\
      Infill/Gen & $3.75$ \\
      Math & $6.87$ \\
      \bottomrule
    \end{tabular}
    \caption{Position Ranks}
    \label{tab:unmatched_errors}
\end{table}

To assess the robustness of the searching module, we tracked the average item position (calculated as page rank $\times 10$) for the matched reward model. Across all sampled categories, the overall average retrieval position was $5.64$ items. As detailed in Table \ref{tab:search_robustness}, all individual sub-categories consistently found the optimal item on the first page, confirming the robustness and high precision of the agent's query generation and search logic.

We further validate the soundness of the framework's design by including a detailed analysis of the generated tool quality (Appendix~\ref{sec:gen_tool_quality}) and an investigation into the reward tool usage within our main experiments (Appendix~\ref{sec:reward_tool_usage}). In summary, code-agents achieve a $94.9\%$ executable rate when utilizing rule/metric-based tools, and we observe a dominant percentage of LLM-based reward tool usage in text-generation tasks.

\subsection{Reward Tool Generation Quality} \label{sec:gen_tool_quality}


We evaluate the quality of reward tools produced by our two agents for tool generation mainly along their construction validity and summarized in Table~\ref{tab:tool_gen_quality}.

\paragraph{Code-agent tools.} 
Across all the training queries, the code agent generated \textbf{118} reward scripts, among which \textbf{112} (\textbf{94.9\%}) were directly executable under our standardized interface.\footnote{Executability is checked by importing the generated function, calling it with a minimal synthetic triplet \texttt{(prompt, candidate, reference)} and verifying a numeric return type without exceptions.}
By type, the set comprises \textbf{102} rule-based functions (\textbf{86.4\%}), \textbf{12} standard metric implementations (\textbf{10.2\%}; e.g., BLEU, METEOR), and \textbf{4} learned-model–based scorers (\textbf{3.4\%}).
Rule-based tools typically encode task-specific verifiable criteria (e.g., numeric-consistency checks for \textsc{GSM8K} or explicit-irrelevance penalties for RLHF-style preference items), while metric-based tools provide length- or n-gram–aware surrogates for general text quality.
We discard learned-model–based proposals from the code agent since they are potential out of memory threats to the deploying server.

\begin{table}[t]
    \centering
    \small
    \begin{tabular}{lc}
    \toprule
    \textbf{Category} & \textbf{Count} \\
    \midrule
    Code-agent scripts (total) & 118 \\
    \quad Executable & 112 (94.9\%)  \\
    \quad Rule-based & 102 (86.4\%) \\
    \quad Metric-based & 12 (10.2\%) \\
    \quad Learned-model–based & 4 (3.4\%) \\
    \midrule
    Web-agent repos (retrieved) & 21 \\
    \quad Deployed & 10 (47.6\%) \\
    \quad Rejected (size) & 2 \\
    \quad Rejected (not classification) & 6 \\
    \quad Rejected (insufficient docs) & 3  \\
    \bottomrule
    \end{tabular}
    \caption{Summary of reward tool generation outcomes.}
    \label{tab:tool_gen_quality}
\end{table}

\paragraph{Web-agent tools.}
The web agent retrieved \textbf{21} candidate repositories from public model hubs (primarily Hugging Face and ModelScope) that matched the predicted task label and satisfied our \emph{reward-model} filter. And the filter eliminates base/instruct/chat/vision models and retains text-classification modeled reward models with download access.
After automatic screening and wrapping, \textbf{10} repositories (\textbf{47.6\%}) were successfully deployed behind a uniform Python API. The remaining \textbf{11} were rejected due to: model size prohibitive for our inference node (\textbf{2}), non–text-classification architectures (\textbf{6}), or insufficient/ambiguous repository documentation for reliable wrapping (\textbf{3}).



The high executability of code-agent tools (\textbf{94.9\%}) and the moderate but reliable deployment rate of web-agent tools (\textbf{47.6\%}) indicate that \method{} can \emph{consistently} materialize task-aligned reward functions across heterogeneous inputs.



\subsection{Reward Tool Usage and Selection Patterns} \label{sec:reward_tool_usage}
Having established that \method{} can reliably generate and deploy reward tools, we next examine \emph{how} these tools are actually invoked during training.
This analysis addresses two questions: (i) which categories of tools dominate in practice, and (ii) how the usage patterns vary with task source and affect the learned policy.

\begin{figure}[t]
  \centering
  \includegraphics[width=0.45\textwidth]{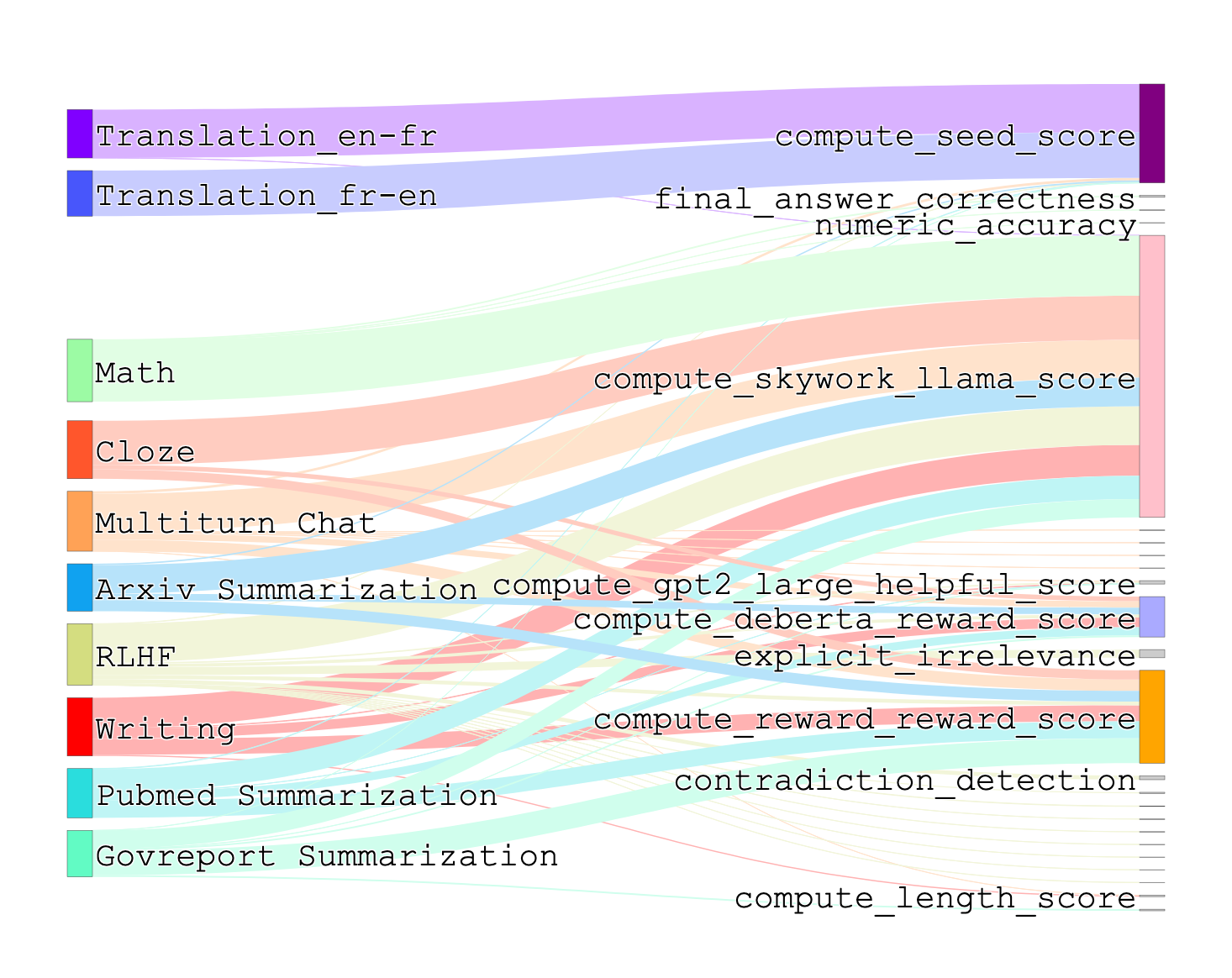}
   \caption{Matching tools with source training dataset distribution.}
  \label{fig:sanky}
\end{figure}

We plot the actual usage of tools by examining the tool matching conditions based on the data source in the training set shown in the Figure~\ref{fig:sanky}.
Across all $8{,}000+$ training samples, the majority of calls are routed to \textbf{LLM- based reward models} (\textbf{96.4\%}), while \textbf{rule- based} and \textbf{metric- based} tools are invoked only sparsely.
The most frequently selected individual model is \texttt{Skywork/ Skywork-Reward-V2-Llama-3.1-8B}, accounting for \textbf{52.5\%} of calls.
A significant proportion of samples fall back to rule-based numeric-consistency checks (``explicit number match'')
On translation tasjs, the web-agent originated \texttt{Seed-X-RM-7B} dominates, capturing cross-lingual adequacy more effectively than generic reward models.

The dominance of LLM-based rewards suggests that, for heterogeneous open-domain training, high-capacity discriminative models remain the most trusted. 
Nevertheless, the occasional use of rule-based checks in math and RLHF tasks demonstrates that \method{} is capable of \emph{combining expert heuristics} when appropriate.
\method{} does not rely on a single global reward model but instead orchestrates a \emph{portfolio} of evaluators aligned with each domain.
As shown in the previous subsection, this diversity translates into smoother advantage estimation and stronger updates during policy optimization.

\subsection{Impact on Advantage Estimation and Policy Learning}

\begin{figure}[t]
    \centering
    \includegraphics[width=1.0\linewidth]{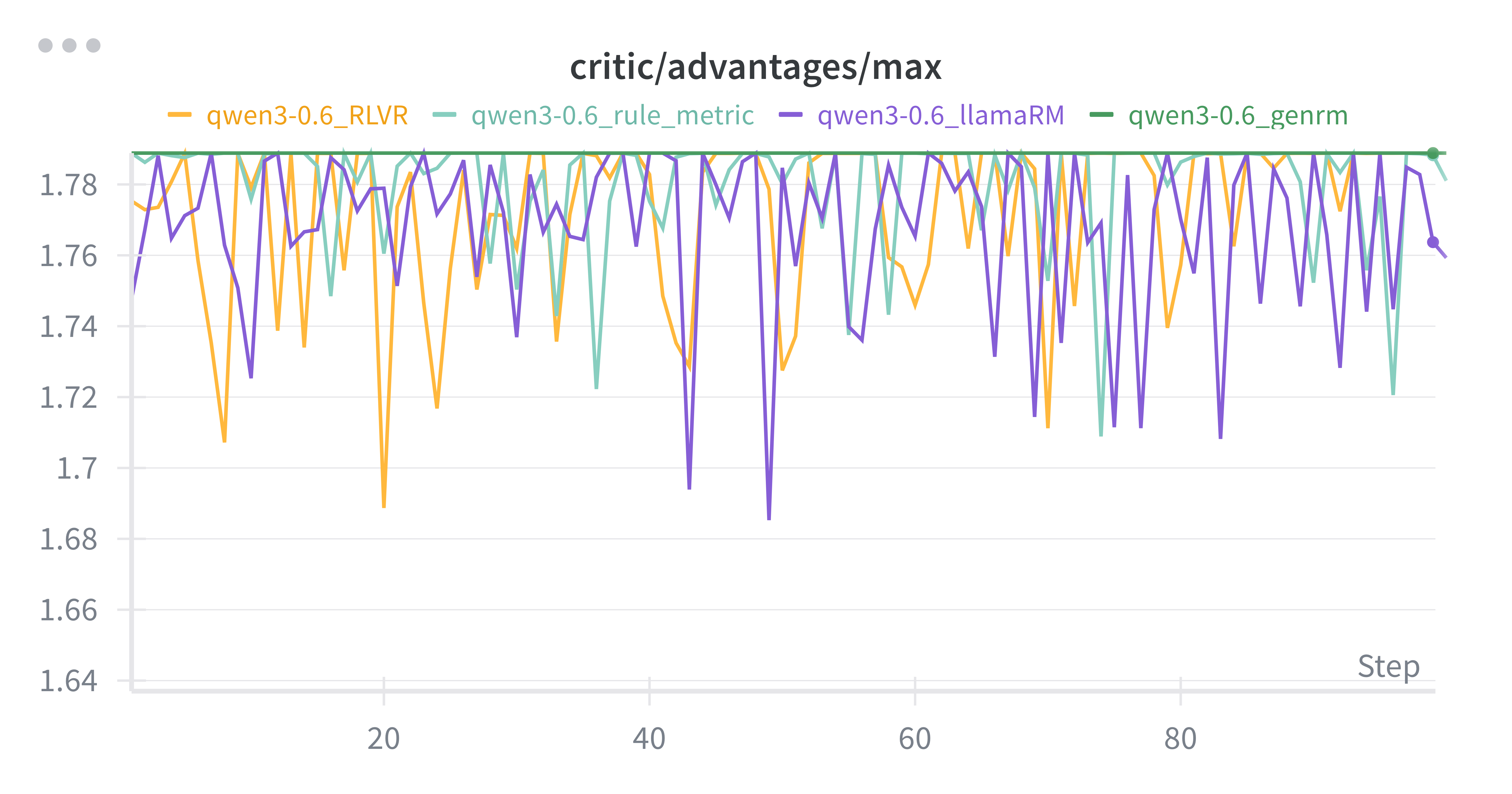}
    \caption{Maximum Advantage Estimations}
    \label{fig:max_adv}
\end{figure}

\begin{figure}[t]
    \centering
    \includegraphics[width=1.0\linewidth]{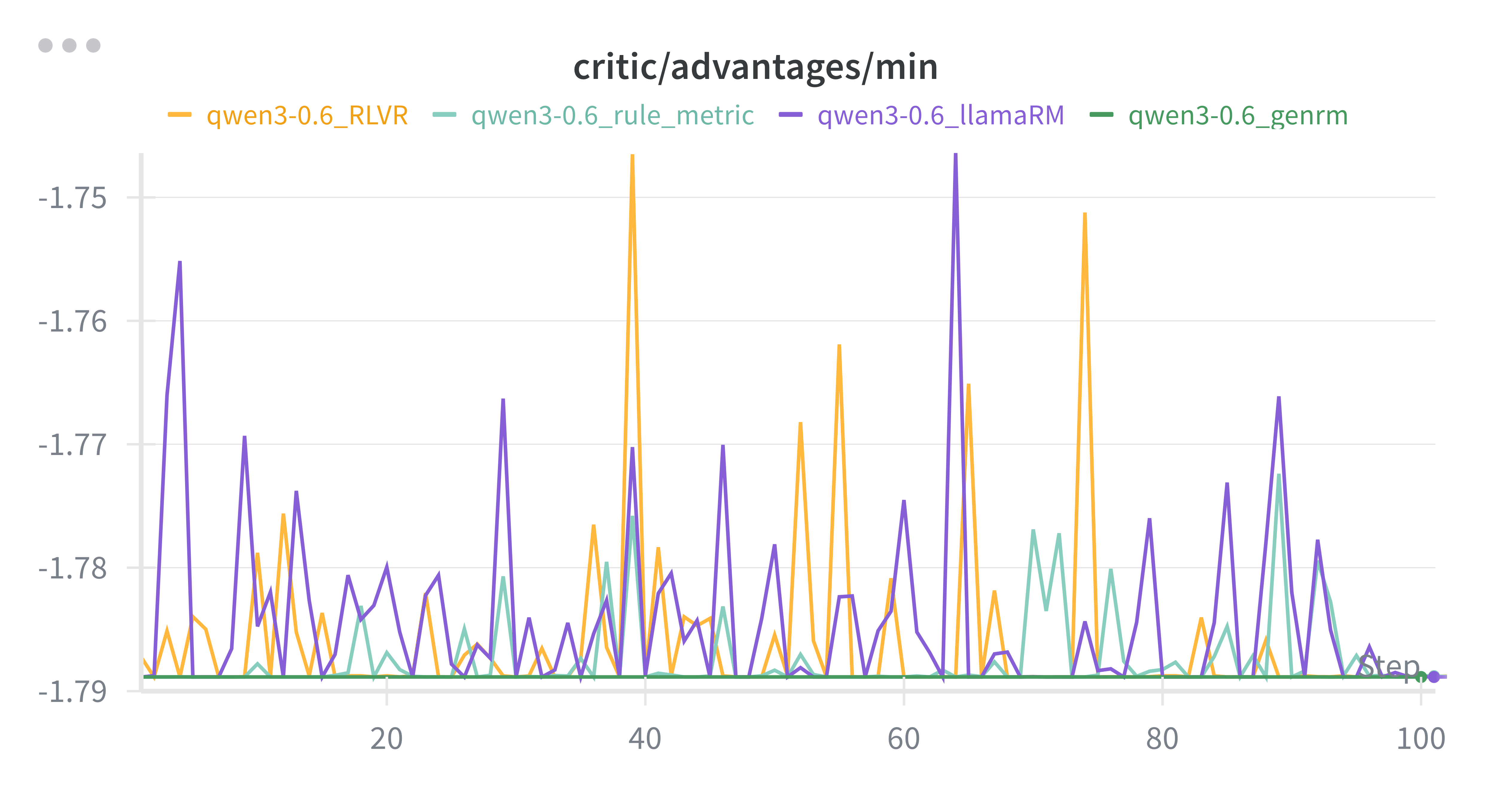}
    \caption{Minimum Advantage Estimations}
    \label{fig:min_adv}
\end{figure}

We examine the records from the Qwen experiments covering Generative RM, method, single generic reward, regarding the estimated min/max of advantage per step (Figure~\ref{fig:max_adv} and Figure~\ref{fig:min_adv}), and calculated the proportion that triggered clipping. Higher rates of being clipped means a higher absolute value of estimated advantage. From the results, for Generative RM, rollouts triggering both upper-clip and under-clip occur in every update step. Compared to single generic reward, \method{} has a significantly higher clipping rate. This is direct evidence that \textbf{methods with better performance tends to estimate larger advantages in absolute values}.

\begin{figure}[t]
  \centering
  \includegraphics[width=1.0\linewidth]{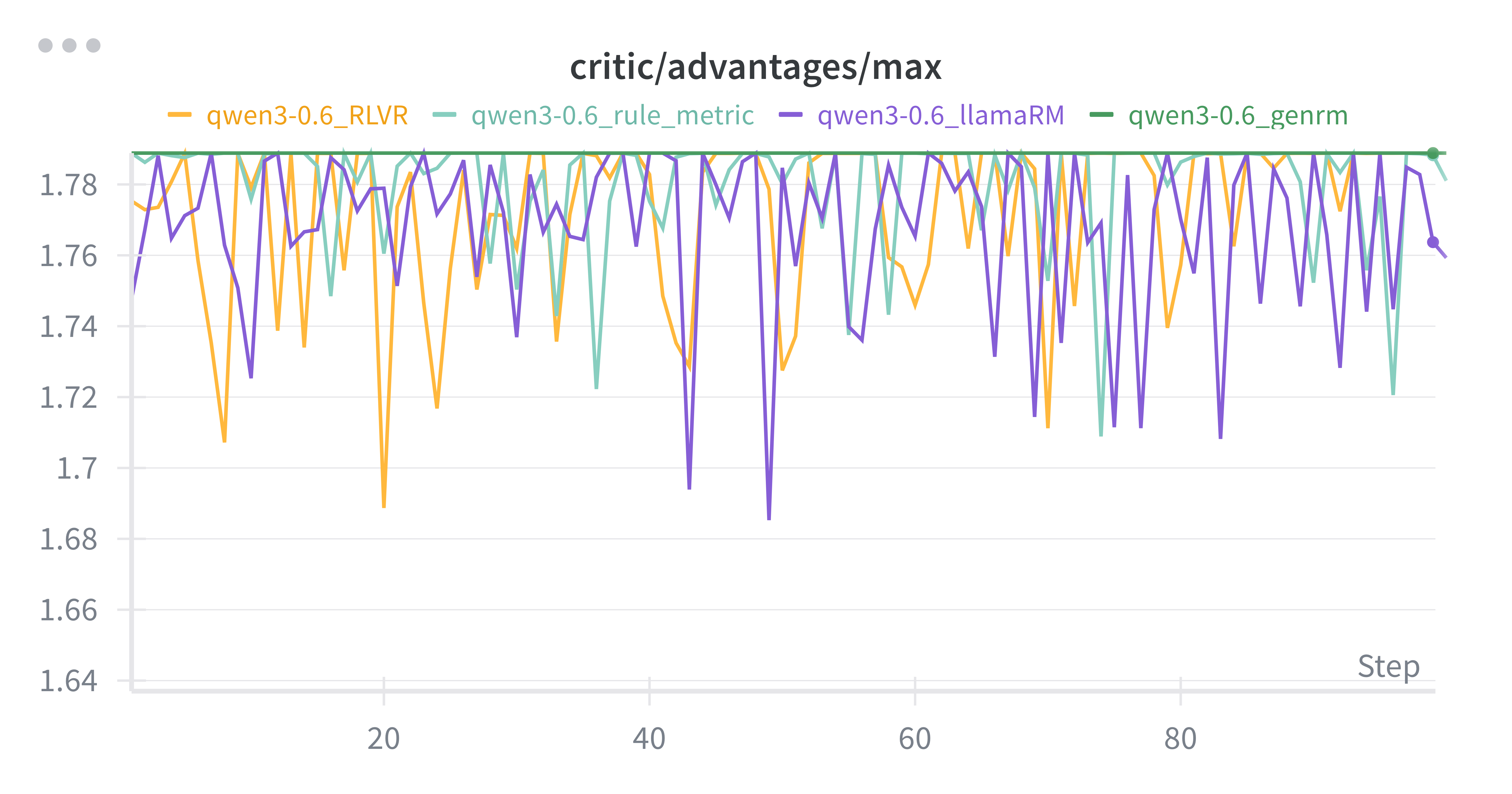}
   \caption{\small An illustration on the sensitivity to extreme values of reward functions.}
  \label{fig:illustrate}
\end{figure}

Return to the discussion of Advantage Estimation \(\hat{A}_{i} = \frac{r_i - \mathrm{mean}(r)}{\mathrm{std}(r)}\). Consider two types of reward functions in Figure~\ref{fig:illustrate}, the blue one is sensitive to extreme values (smaller variance) while the orange one is evenly modeled (higher variance). 
Assuming uniform roll-out sampling, a higher value of \(\hat{A}_i\) suggests that the underlying reward function resembles \textbf{the sensitive type} (blue line). Therefore, extreme values (maximum/minimum) are divided by a smaller variance, resulting in a more frequent reaching of the clip threshold. This is expected for policy optimization that more weights should be transferred to these deviated rolls, as part of exploration-exploitation balance.

\end{document}